\documentclass[conference]{IEEEtran}
\usepackage{cite}
\usepackage[pdftex]{graphicx}
\usepackage[cmex10]{amsmath}
\usepackage{amssymb}

\usepackage{times}
\usepackage{array}
\usepackage[tight,footnotesize]{subfigure}
\usepackage[font=footnotesize]{subfig}
\usepackage{graphicx}
\usepackage{mathtools}
\usepackage{microtype}
\usepackage{amsmath}
\def\union{{\cup}}

\usepackage{alltt}
\usepackage{shadethm}
\newshadetheorem{comment}{Comment}

\def\boxx{{\vcenter{\vbox{\hrule height.3pt
          \hbox{\vrule width.3pt height6pt
          \kern6pt\vrule width.3pt}\hrule height.3pt}}\;}}

\def\impos{{\;\vcenter{\hbox{\rule{5mm}{0.2mm}}} \vcenter{\hbox{\rule{1.5mm}{1.5mm}}} \;}}

\def\lrarrow{\leftrightarrow \kern-8pt \rightarrow}

\def\2{\frac{1}{2}}

\def\beq{\begin{eqnarray}}
\def\eeq{\end{eqnarray}}
\def\2{\frac{1}{2}}

\newtheorem{definition}{Definition}

\newtheorem{hype}{Hypothesis}

\def\lrarrow{\leftrightarrow \kern-8pt \rightarrow}

\def\frightarrow{\rightarrow \kern-11pt /~~}
\def\reducesto{\simeq \kern -3pt >}
\def\intersection{\cap}

\usepackage{fancyhdr}					
\begin{document}
\newcommand{\strust}[1]{\stackrel{\tau:#1}{\longrightarrow}}
\newcommand{\trust}[1]{\stackrel{#1}{{\rm\bf ~Trusts~}}}
\newcommand{\promise}[1]{\xrightarrow{#1}}
\newcommand{\revpromise}[1]{\xleftarrow{#1} }
\newcommand{\assoc}[1]{{\xrightharpoondown{#1}} }
\newcommand{\rassoc}[1]{{\xleftharpoondown{#1}} }
\newcommand{\imposition}[1]{\stackrel{#1}{\impos}}
\newcommand{\scopepromise}[2]{\xrightarrow[#2]{#1}}
\newcommand{\handshake}[1]{\xleftrightarrow{#1} \kern-8pt \xrightarrow{} }
\newcommand{\cpromise}[1]{\stackrel{#1}{\frightarrow}}
\newcommand{\policy}{\stackrel{P}{\equiv}}
\newcommand{\field}[1]{\mathbf{#1}}
\newcommand{\bundle}[1]{\stackrel{#1}{\Longrightarrow}}

\title{Testing the Quantitative Spacetime Hypothesis using Artificial Narrative Comprehension (II)\\~\\\normalsize Establishing the Geometry of Invariant Concepts, Themes, and Namespaces}

\author{Mark Burgess}
\maketitle
\IEEEpeerreviewmaketitle

\renewcommand{\arraystretch}{1.4}

\begin{abstract}
  Given a pool of observations selected from a sensor stream, input
  data can be robustly represented, via a multiscale process, in terms
  of invariant concepts, and themes. Applying this to episodic natural
  language data, one may obtain a graph geometry associated with the
  decomposition, which is a direct encoding of spacetime relationships for
  the events.

  This study contributes to an ongoing application of the Semantic
  Spacetime Hypothesis, and demonstrates the unsupervised analysis of
  narrative texts using inexpensive computational methods without
  knowledge of linguistics. Data streams are parsed and fractionated
  into small constituents, by multiscale interferometry, in the manner
  of bioinformatic analysis.  Fragments may then be recombined to
  construct original sensory episodes---or form new narratives by a
  chemistry of association and pattern reconstruction, based only on
  the four fundamental spacetime relationships. 

  There is a straightforward correspondence between bioinformatic
  processes and this cognitive representation of natural language.  Features
  identifiable as `concepts' and `narrative themes' span three main
  scales (micro, meso, and macro). Fragments of the input act as
  symbols in a hierarchy of alphabets that define new effective
  languages at each scale.
\end{abstract}

\hyphenation{similarity}
\tableofcontents


\section{Introduction} 

Can we demystify basic processes of cognition, in real and Artificial
Intelligence research, by appealing to some general principles of
scale and consistency?  Modern AI research often builds on the
application of black box technologies as tools, and models do little
to shed light on how artificial cognition might naturally be
considered part of a larger class of processes appearing over multiple
scales.  The Spacetime Hypothesis, proposed
earlier\cite{spacetime1,spacetime2,spacetime3,smartspacetime} offers a
simple proposition: namely that spacetime processes must underpin all
aspects of cognition. The goal in this series of papers is to test the
(rather large) implications of that hypothesis explicitly with a
working model\footnote{A number of authors has attempted to formulate
  speculative or toy models of consciousness on a philosophical level,
  based variously on ideas from Information Theory and Quantum
  Mechanics, but these are hard to take
  seriously\cite{tononi,tegmark}.  To join those ranks is not the
  intent of this work.}. The scope of the problem is large, so it can't
be covered in a single work.

Measurement scales (engineering dimensions) are the basis of all
descriptions of natural processes in physics and chemistry, yet they
are often deliberately eliminated in statistical studies. The appeal
of probabilistic methods, to elicit `universal characteristics' and
scale-invariance, can bring confusion rather than clarity. For example,
in language studies relating to the present work, the works of Zipf
and Mandelbrot\cite{zipf1,mandelbrot1,zipflanguage} famously remarked
upon scale-invariant distributions as properties of language. However,
such procedures purposely eliminate an important source of
information: cross-dimensional scales that characterize the relative
interactions between the object of study and their environment.  The
Spacetime Hypothesis contends that we have to return to a natural
scale analysis (not a scale-free one) to understand
phenomena\cite{spacetime1,spacetime2,spacetime3,smartspacetime}.  The
treatment of scales has a long history in physics\cite{scale1,scale2}.

In natural language analysis, linguistics also tend to forego
quantitative scales, preferring to focus attention on the
reduction of functional (semantic) elements, i.e. components of
grammar.  Once a mindset for grammatical thinking has been
established, it's hard to disregard one's own knowledge of language in
the approach.  Here, the Spacetime Hypothesis takes a deliberately
different approach, more reminiscent of biological analysis. We may
retain semantics, in the form of distinct quasi-symbols, to look at
raw pattern fragments, and the commonality of such accumulated
ingredients (a procedure one might call symbolic
interferometry\footnote{This approach will inevitably entail
  limitations, including those noted in the {\em interferometric
    equivalence principle}\cite{mandelwolf}---but these are precisely
  the limitations we have to confront in understanding how to
  bootstrap meaning from sensory input, and---by implication---in
  language too.}).

The successes of Artificial Neural Networks (ANN), or processes
inspired by neurobiology, have left many willing to forego a causal
understanding of recognition methods, attributing successes to almost
mystical properties of specific Machine Learning apparatuses.  This
has led some to a premature rejection of `symbolic approaches' to
Artificial Reasoning\footnote{See for example the review of the state
  of affairs in \cite{marcus1}}.  However, there remains a gulf of
understanding between statistical inference methods and the origin of
logical reasoning\cite{pearl1,pearl2,pearl3,pearl4}.  This work
illustrates one way in which the two descriptions might plausibly come
together---by understanding the scaling of symbolic representations.

The paper follows directly from the prior study in \cite{cognitive4}
(hereafter referred to as paper 1), and applies an approach that
revisits ideas developed with A. Couch in
\cite{inferences,stories}.  In paper 1, natural language texts (episodes of narrative) were
used as a data source, ignoring a linguistic understanding of their
content. Data were simply assumed to express `spacetime phenomena' from
which one then tried to extract meaningful structures, building on the hypothesis
that spacetime patterns determine significance.  From those
structures, meaningful patterns could be identified.
Paper 1 showed that a principle of fractionation was important to
identify invariants and define scales inherent in the spacetime
structure of the text.  

In this sequel, two questions are addressed.  Given the basic
constitution of pattern fragments from paper 1, which converts input
data into an alphabet of new effective symbols, 
\begin{itemize}
\item How should one
organize fragments into a knowledge representation that retains their
spacetime relationships to one another?  
\item Moreover, how do the data
form representations of concepts (independent of the input language)
such that a cognitive agent could tell its own stories, on a new
level, based on the emergent geometry of its memory representation?
\end{itemize}

Following the results of \cite{cognitive4}, we have the simple
understanding of how quantitative measures behave within streams of
symbolic data---in a way that can presumably be extended to
non-digital patterns.  The next step is to study whether or not we can
order, rank, and extract meaningful geometry for reasoning about the
event patterns, their fractions, and aggregations representing
context, all without any linguistic understanding. This is the essence
of Automated Reasoning (AR)\footnote{The approach used here shares a few similarities with machine-learning
approaches to mining ontologies from text data (see for instance
\cite{fanizzi2,fanizzi1}), but in practice the approach is
deliberately less sophisticated, and we shall not assume annotated
logical properties, just as we assume nothing about grammatical decomposition.}.

Once a stable geometry has been established, a key test is whether we
can generate narratives from the data whose translations would be
acceptable interpretations of the story.  The procedure can be
compared to the Turing Test, which is not a quantitative result but
rather a qualitative assessment of whether the encoding functions in a
believable way.  Crucially, we shouldn't expect a higher standard of
an artificial system than we would of humans.  Even `intelligent
humans' speak apparent nonsense at times for various reasons.  The
results are partially convincing in principle, but focus on the
mesoscopic scale. Results points to the need for further study
on the macroscopic interactions.

\section{Projecting narrative into Semantic Spacetime}

The semantic spacetime model predicts the existence of a
graphical representation for processes, based on four elementary types of
relationship between nodes in an agent model.  Although, as humans, we
read narrative in a particular manner---by convention---there are
other ways to read it too: browsing, indirection or jumping into the
middle via an index, etc. This is better represented by a graph
structure than by a stream, because time is what is experienced by a
process of observation, selected from a source process,
not determined by the source alone.

\subsection{Definitions and hypotheses concerning the construction of a causal graph from fragments and their co-activation contexts} 

Discussing languages and meta-languages within data, based on natural language,
potentially leads to some confusion of terms.
To avoid some semantic muddle, let's define the following nomenclature:

\begin{figure}[ht]
\begin{center}
\includegraphics[width=4.5cm]{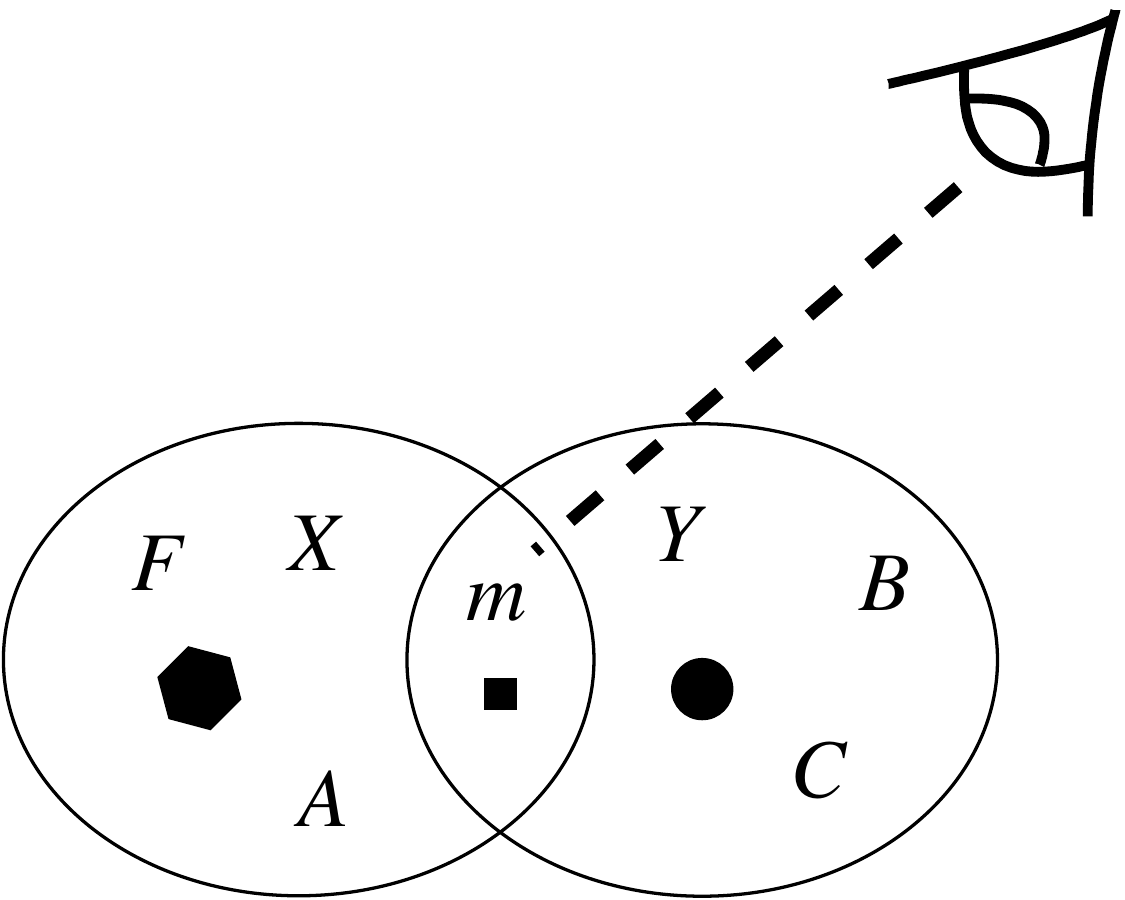}
\caption{\small Context (right) is used by the observer to measure similarity.
Those symbols that co-exist in the source and in the running context become
`co-active' and indicate the degree of overlap.\label{overlap_context}}
\end{center}
\end{figure}

\begin{definition}[Input language]
The stream of English symbols and words parsable by the sensory receiver (depicted in
figure \ref{overlap_context}). The stream is chopped into fragments $\phi_n$ which
retain the original language symbology.
\end{definition}
Note carefully, although human language comprises the content of the stream
of patterns perceived by the sensor, our ability to read it is of no
consequence to the processing: it plays no role in the method of
construction of events, hubs, and regions. Our knowledge of the
language gives us a privileged advantage in assessing the fragments at
certain stages of the analysis, but no grammatical criteria are used
in this work, except for spaces and full stops (periods).  The human language (English), used in this study, is
considered only as an encoding on the level of patterns. It will be
for us, as privileged observers, to comment on any correlation between
them at a later stage of the argument.

\begin{definition}[Concept Language]
The elements in a graph representation of the input
data condense into a new set of invariants. These, together
with the links that express relations between them,
comprise a new language---which is the language of the
Artificial Cognitive Agent.
\end{definition}

The concept language's symbols and phrases are built as a dynamic
process of discovery during the processing of the stream.  The
eventual stabilization of any dynamical structure generally requires
both sufficient diversity and sufficient bulk or `mass' to build a
complete picture.
 That occurs both via the persistence of fragments
representing contextual fragments stored in hubs, and by the
subsequent super-aggregation resulting from the joining of hubs into
larger regions.  We might therefore posit the following, based on
simple scaling arguments:

\begin{hype}[Invariant local patterns are concepts]
  Patterns that recur from stimulus by the environment are the source
  of concepts, at each scale. The different scales refer to different
language representations.
\end{hype}
The case for this can be substantially confirmed.

\begin{hype}[Observations are input strings]
Strings of the input language represent behaviours in the
exterior world of the cognitive agent that express causal and
spatial information.
\end{hype}
This is more of an axiom, but as a source of complete, no other
sources of data are needed in the study.

\begin{hype}[Statements are graph trajectories]
  Paths through the graph of relational `promises' represent the
  cognitive agent's process of reasoning, i.e. formulating statements
  about the world perceived by the cognitive agent in the concept
  language.
\end{hype}
This proposal is harder to determine unequivocally, but the results
point to the reasonableness of the assumption.  The paths might be
very long, especially when crossing episodes transversely.  The
challenge is that the span of scales involved in conceptualization is
so vast that this preliminary work can't capture every angle or cross
off every objection. Nonetheless, the manifesto seems promising.

It's worth repeating once again that the language referred to in this
second hypothesis should not be confused with the symbolic language
that happens to be the source of data in this study. The two languages
are quite independent, but the potential for confusion is strong
because we are using one language to study how another (potentially
like it) could emerge by cognitive scaling. To make matters even more
confusing, we must eventually use the input language to explain the
symbols of the derived language too (because that's the language you
are reading now).  Finally, the summation of these ideas:

\begin{hype}[Narrative]
Once a cognitive system has learned by merging several narratives, it
has the potential to tell new stories---by combining inferences across
the connected network based on its long term memory.
\end{hype}
This too seems promising, and this work will help to 
clarify more convincingly once we can generalize the approach
to other kinds of data.

\subsection{The four semantic relations of spacetime}

Although the specifics of semantics relationships may entail
a wide variety of subtle interpretations, or `subtypes', the
Spacetime Hypothesis proposes that these must all belong to four basic
spacetime types: 
\begin{itemize}
\item {\sc FOLLOWS}: Events follow one another in process time $t$,
  according to some partial order relation `$>$'.  The narrative process
  has a partial order which can be retained from events, to link up
  episodic events into a chain. Sentence evens may thus promise to
  follow one another, e.g.

\beq
S_t \promise{+(t' > t)} S_{t'}.
\eeq
This relation is a strong binding interaction.

\item {\sc CONTAINS}: Collections of agents can be considered parts of a larger whole.
Thus a collection contains member agents, which allows scaling of identity.
A collection of agents can be unified by connecting each member to a central hub agent.
Hubs $H_i$ promise to represent clusters of sentence events $S_t$, and sentence events contain all possible
fragments within them.
\beq
H_i &\promise{+S_t}& O,\\
S_t &\promise{+\phi_n}& O,
\ldots
\eeq

\item {\sc EXPRESSES}: Invariant patterns express information which distinguishes them from one another.
The identity or proper names of agents are thus `expressed' as `scalar promises', or self-properties,
On a larger scale, the same is true of aggregations of agents, acting a superagents. Sentences can express
fragments:
\beq
S_t \promise{+\{\phi_n\}} O.
\eeq
Then by implication, hubs also promise the sum fragments that compose them
\beq
H_i \promise{+\{\phi_n\}} O.
\eeq

\begin{figure}[ht]
\begin{center}
\includegraphics[width=6.5cm]{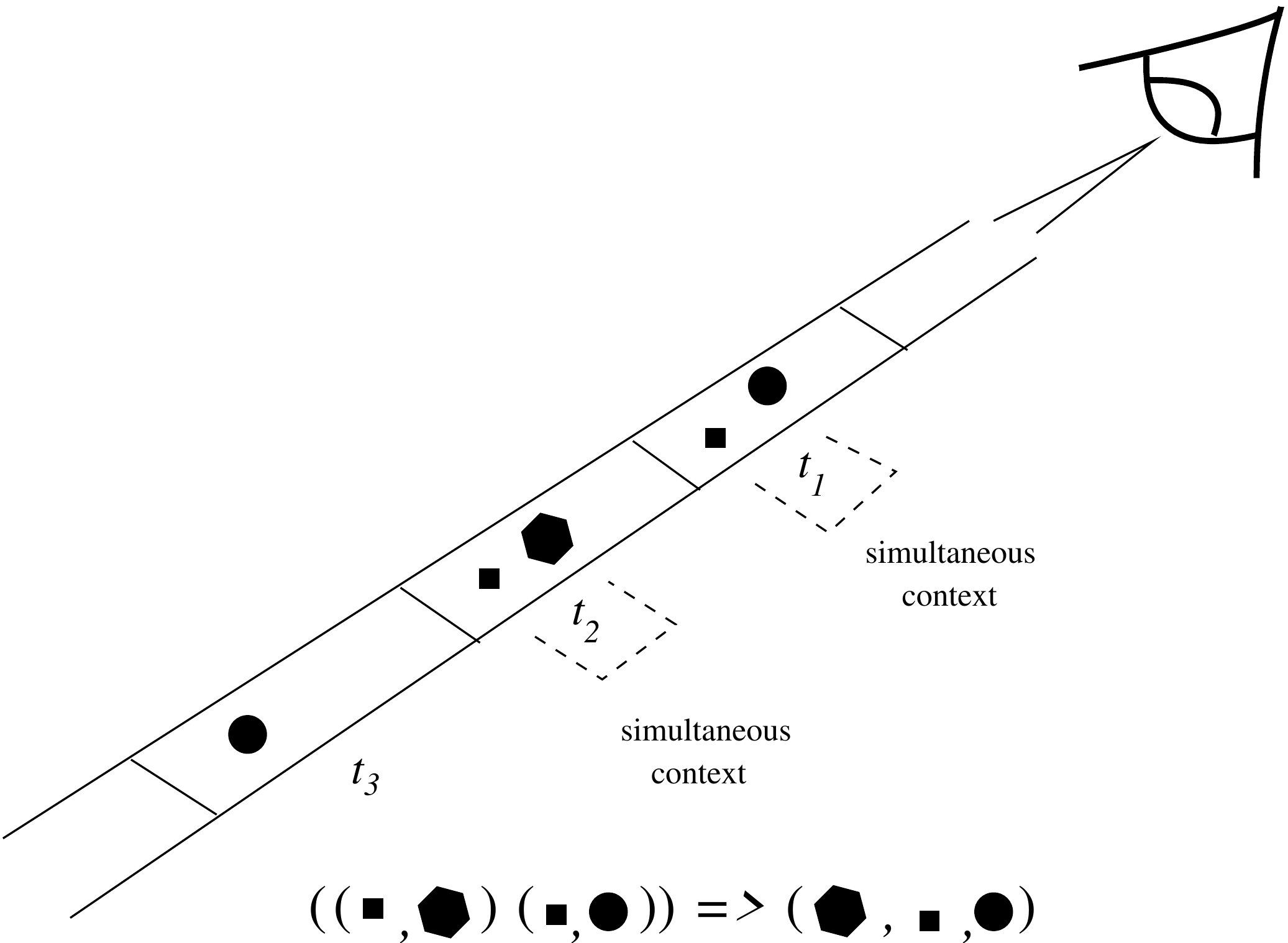}
\caption{\small The longitudinal data stream is coarse-grained into
  legs and fractionated into sentence events $S_t$ and n-phrases
  $\phi_n$, which form hubs $H_i$ by co-activation of phrases in
  proper time $t$. Each fractional $\phi_n$ can be associated
  (transversely) with the hubs it belongs to by post-processing, and
  events follow one another. The overlap between sets of fractional
  parts determine their `closeness' at each level.  This post
  processing is the only way to associate similar ideas out of band,
  by smoothing over the data representations without being tied to the
  sequence driven by sensory input.\label{coactivation}}
\end{center}
\end{figure}

\item {\sc SIMILAR TO (NEAR)}: Agents (superagents) that make collections of 
promises can be compared on the basis of their common promises; thus one may attribute a degree of similarity between
them. If they express precisely the same $\phi_n$, they are maximally similar
(proximal, or close together -- see figure \ref{coactivation}). If they don't overlap at all, then they are disconnected silos.
Closeness, in fragment space, therefore comes from the interferometry of fractional sets.
A few random overlaps of fragments may lead to remote or weak connections (far apart means few similarities),
and, as we'll see below' a cognitive agent which excludes those beyond a certain horizon
will be more successful in separating concepts than one which eagerly relates all things.
The degree of overlap between hubs, defined as the coincident members:
\beq
d_{ij} = \frac{2(H_i \intersection H_j)}{(H_i \union H_j)} \times 100\%
\eeq
The proximity relation is a weak binding interaction.

\end{itemize}
These types have been used earlier in
\cite{spacetime3,cognitive,koaljahistory,observability}.
Within the graph representation, the relations are represented by links or edges
of the graph\cite{schaumgraph,west1,berge1}. Within expressed quantities, they are represented
by direct adjacency of symbols as strings.

\begin{figure*}[ht]
\begin{center}
\includegraphics[width=13cm]{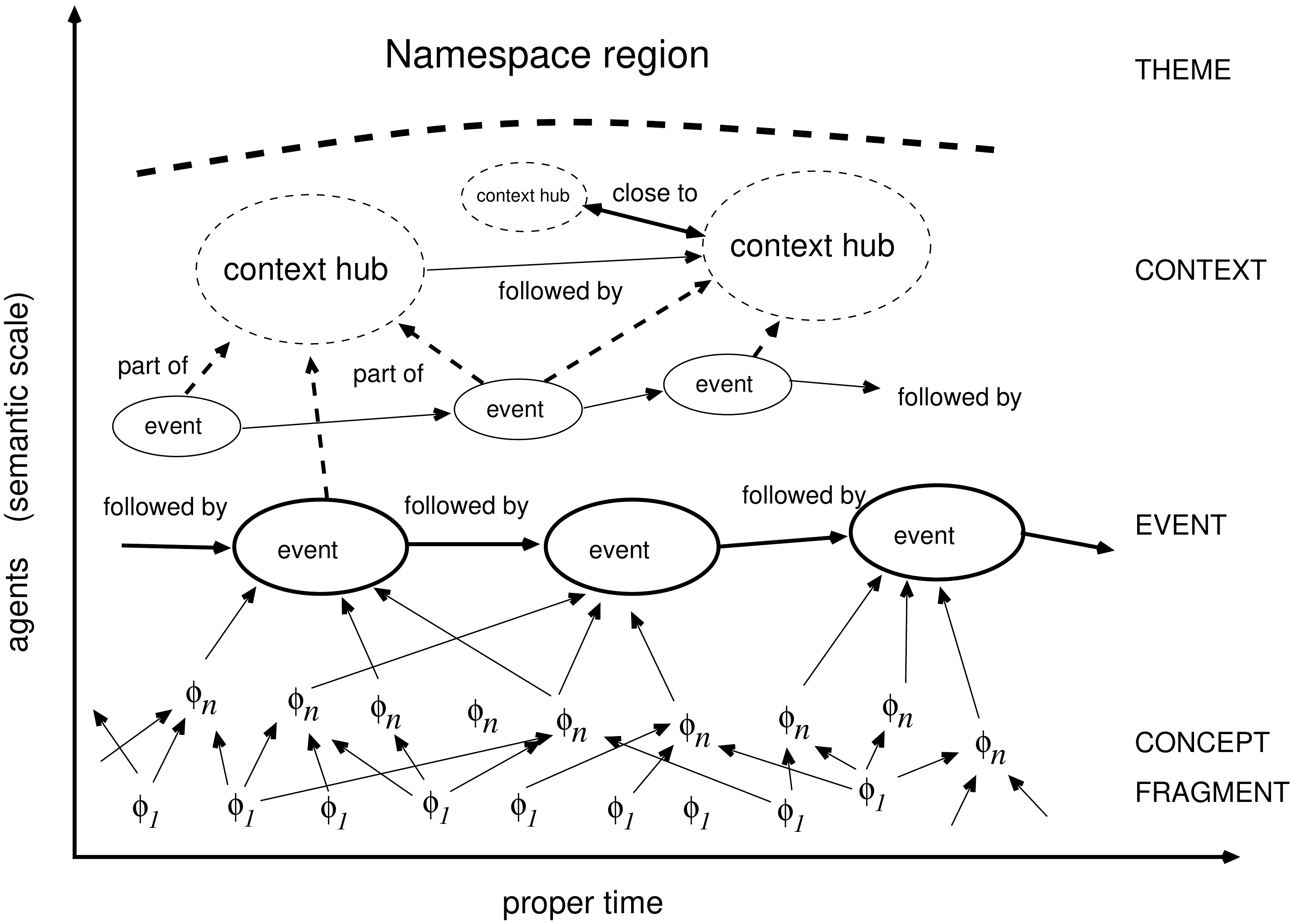}
\caption{\small Spacetime structure is constructed semantically on four
  kinds of relation. The privileged axis of this diagram lies around
  event chains.  These express phrase fragments $\phi_n$ downwards,
  and are aggregated within concepts upwards. The accretion of concepts from
events may lead to significant overlap, which places some concepts closer
to others by virtue of similarity of constituents. This is a form of
detailed co-activation in the network of the lower layers.
Concepts are an accumulation of fragments, accumulated through the
causal process of sensory cognition. Events may eventually be generated
from combinations of concepts or phrases, and treated recursively to the
same process as in paper I.\label{spacetime}}
\end{center}
\end{figure*}

\subsection{Fractionation of sequential data}

To render highly specific combinatoric sequences into a comparable
form, one may fractionate them into mixtures of their smallest constituents
and measure their spectra. Thus each sentence is broken up into parts of
different word length $\phi_n$ (see paper 1).

In paper 1, single narrative (document) sources were fed into a
preprocessor, which chopped up the stream into sentences, and chopped
each sentence into $n$-phrases $\phi_n$, i.e. sequences of $n$ words
bounded by the sentence (for $n=1,\ldots,6$). In each `leg' (or
quasi-paragraph) of a stream the statistical characteristics of the
phrases are used to rank their importance (see figure
\ref{coactivation}). Any significant changes in the spacetime measures
of the sampling process may lead to a state of greater `attention' or
higher sampling, otherwise a low level of sampling is maintained.

Overall about one part in a hundred of the stream was typically
extracted, based on importance. This is an arbitrary choice taken in
paper 1 and continued here for consistency. In a more effective
application, the density should perhaps be higher.  The selections retained
whole unedited sentences as hubs for the member fragments, based on
their $\phi_n$ importance scores.

\begin{figure}[ht]
\begin{center}
\includegraphics[width=8.5cm]{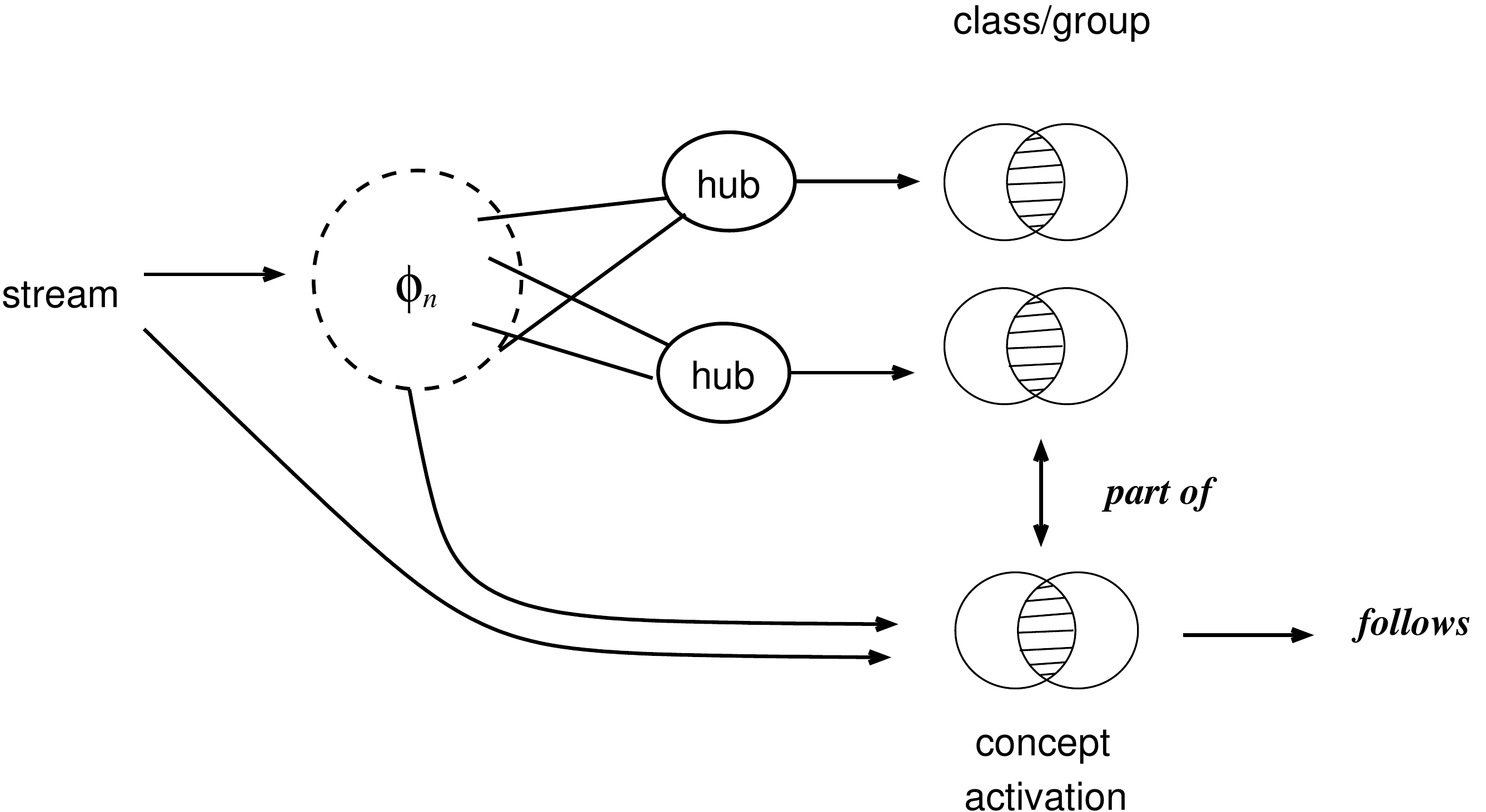}
\caption{\small The geometry of the data stream becomes
  two-dimensional (space and time) as it is fractionated into
  n-phrases $\phi_n$, which form hubs by co-activation of phrases,
  (longitudinally or `follows') in proper time. Each fractional
  $\phi_n$ can be associated (transversely or `part of') with the hubs
  it belongs to, and events follow one another. The overlap between
  sets of fractional parts determine their `closeness' at each
  level.\label{coactivation2}}
\end{center}
\end{figure}

The sequence of transformations is thus (figure \ref{coactivation2}):
\begin{enumerate}
\item Sentence selection of events by importance score: text $\rightarrow S_t$
\item Fractionation of all sentences into $n$-phrases: $S_t \rightarrow \phi_n$ for rolling context.
\item Assembly of $\phi_n$ into sets, joined to a hub representing a moment of context.
\item Selected sentences are linked to the nearest rolling context hub.
\item Post processing of context hubs to look for similarities based on mutual information.
\end{enumerate}

All the learning here may be characterized as unsupervised, and
happens by realtime assessment of fragments $\phi_n$.  All data are
reset at the start of each narrative experiment, unless the experiment
concerns a merging of the narratives. Fragments are
forgotten at a controlled rate to maintain a dynamic `pressure' or
equilibrium to resist random selection. The forgetting rate was tuned
so as to never forget the most common words (`of', `the', etc) that
punctuate and glue more significant phrases together.

\subsection{Role of the observer and running context}

Machine learning models typically construct classical `God's-eye view'
models of the world, in the Cartesian-Newtonian tradition: a single
truth for all in the system.  The Spacetime Hypothesis, by contrast,
automatically leads to a Local Observer View of data, derived from its
promise theoretic origins\cite{promisebook}, meaning that every
observer potentially ends up with a different (relativistic) picture
of the world, depending on the sample and order of its experiences.

As each observer receives a partially-ordered stream of input data, it
uses its short-term memory to count what it sees.  As data
are received from different vantage points, the order and quality of
information may vary. Some information can be blocked from
propagating.  These differences influence the subsequent ranking of
observations within the cognitive process; that, in turn, determines
which fragments to keep and which to discard as noise (see paper 1).

Pattern fragments of data $\phi_n$ are kept in a running `buffer', by a cognitive agent,
and these characterize the here-and-now.  We associate this buffer with the agent's
{\em context}, and this is used in the determination of `co-activation' (see
figure \ref{coactivation}).

The significance of this bank of running fragments (called the agent's
running context) should not be underestimated.  In practice, this is
one of the few available criteria available to an agent for the discrimination
of pathways in a reasoning process. In other words, all
decisions and linkages that transcend simple episodic recall will have
to rely on this running context to make leaps of thought.

\subsection{Running context}

Context appears in three distinct roles in this work:
\begin{enumerate}
\item A running context is accumulated over the different `legs' of
  sensory input, and form the effective information of the leg. This
  context acts as a parameter in the ranking of events during the encoding of memory.

\item Context is effectively cached like a sample of the environment
  in event hubs, both as unifying addressing construct, and as as a
  quick lookup route to adjacent similarity of ideas.  Because context
  is quasi-symbolic, it can be matched, in the manner of mutual
  information between any two hubs, implying a metric distance between
  them.  Hubs that are close together form effective regions over
  short range interaction scales (figure \ref{overlap_context}).

\item Finally, when an agent's thinking processes are not dominated by
  new input, it derives principally from the exploration of memory
  paths. The running context in 1. is then modified by the fragments
  expressed by already-known concepts visited along current search
  patterns. Such context can be matched for similarity to select
  likely relevant pathways in the memory geometry.
\end{enumerate}
At every stage, the running context cache---like a mixed chemical sample or
`primordial soup' of fragments---is the effective selector of
reasoning pathways in the graph representation, somewhat analogous to
chemical spectroscopy of the fragments.

\subsection{Implicit geometry}\label{geom}

The interplay between order and scale leads to geometrical notions
that we shall exploit in encoding information for analysis, reconstruction,
and later recombination.

A narrative begins with a collection of sentences, which are selected
from the raw narrative by an assessment of their significance, then
fragmented into components $\phi_n$\cite{cognitive4}.  They are
located within successive `legs' of the stream's journey.  Legs are a
simple quantitative proxy for context. More important is the semantic
measure of context, which is the rolling collection of most
significant $\phi_n$, given a constant forget rate. Notice how the
forget rate becomes the effective reference calibrator for the time coordinate, and
for context changes. This will play into the rate at which clusters
can form into proto-concepts.  As in all interactions, there are (+)
and (-) promise components\cite{promisebook}.

\begin{itemize}
\item Promised invariants (+), repeating patterns that present at the input.
\item How these are received and classified (-).
\end{itemize}

Invariants act as agents (on a new level) in the memory representation of the cognitive agent.
For the remainder of the paper, agents refer to the space of knowledge representation.
Some agents are in non-ordered phase (liquid) while others are
rigidly ordered (solid phase).
The order of words $\phi_1$ is retained within larger fragments:
\beq
\phi_2 = \phi_1 \text{ followed by } \phi_1'
\eeq
Similarly, at the scale of events, the order of sentences is retained, by linking with `follows'
promises, as these are assumed to capture episodic summaries, with at
least partially causal order. In practice, at the retention rate of
one sentence in two hundred, the order of events is rarely very
significant; however, if we changed the sampling density to a much
higher level, we have to assume that it would be. The order of hub contexts
may also be retained as superagent, even while the fragments that are `contained'
within are not ordered on the interior.

To codify and reconstruct a facsimile of the narrative, the approach
is thus to use these four spacetime semantic relationships to build a
knowledge representation based on a semantic spacetime promise graph,
identified in \cite{spacetime3,cognitive}.  We take the fragments of
sentences $\phi_n$, and connect them as follows These are applied as
in figure\ref{spacetime}.

\begin{itemize}
\item Sentences become agents. They express their content as an atomic unit of narrative.

\item The contents of fragments (which have the status of symbols, i.e. an atomic instance of a
  proper name) are expressed by each fragment. 

\item By counting sentences as units of `proper time', a finite buffer
  size aggregates sentences into coarse grains of narrative progress called `legs'.
  Sentences that score above a certain threshold for acceptance become
  aggregated into grains, and promise to be part of a superagent
  called a hub (denoted $H_i$).  A hub therefore contains sentences,
  and each sentence expresses multiple $\phi_n$ fragment attributes.
  Each hub therefore expresses the sum of those attributes too---which
  summarizes a {\em context}.

\item Sentences express microscopic ordered combinations of words and
  phrases $\phi_n$.  The meaningful sentences $S_t$ promise to
  `follow' each other in the proper time order of the narrative
  (labelled $t$).  Hubs follow one another too when derived from the
  sentence event order.

\item Fragments $\phi_n$ are contained by larger sequences, which are
  `contained' by sentence agents $S_t$, which are contained by hubs in
  their respective legs.

\item The function of smallest fragments is to
  match with similar patterns in other sentence agents. The function
  of longer fragments is to encode uniqueness.  Beyond $n=3$,
  fragments rarely recur\cite{cognitive4}.  These $\phi_n$ become the
  bodies of ($\pm$) promises to offer and accept information, much as
  molecular sequences allow binding between cells or polymers.
\end{itemize}

 To eliminate the constraints of order, but retain components in a
  `mixture' or `solution' form, the partial fragments are aggregated
  into the names expressed by hub structures. The names are thus non-causal,
  non-directional promises.

\begin{figure}[t]
\begin{center}
\includegraphics[width=9cm]{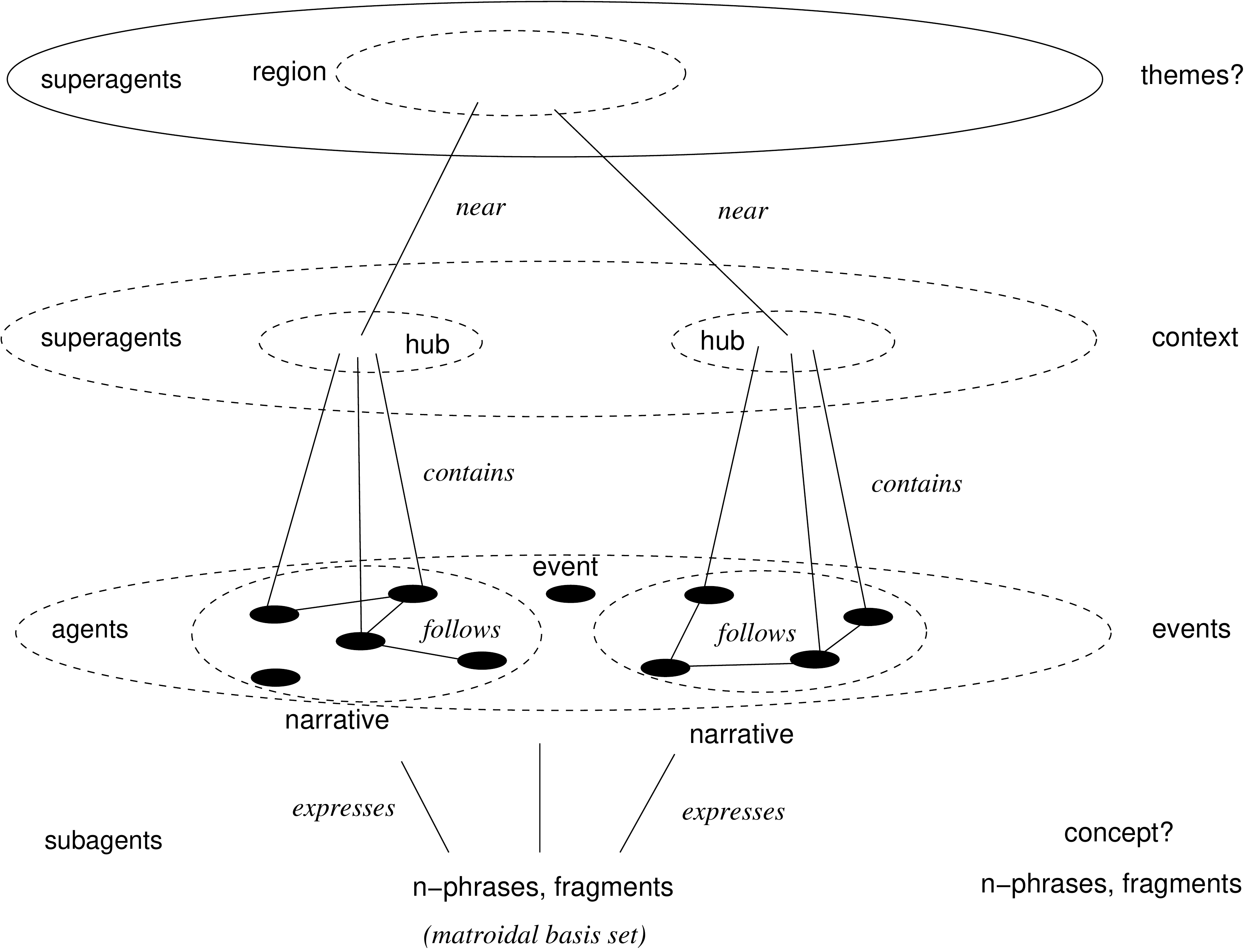}
\caption{\small The hierarchy of graphical agents formed by the spacetime hypothesis.
A redrawing of figure \ref{spacetime}. Where do concepts and themes occur in this scale
diagram? Themes can be associated with patterns of the input language, and concepts of the input
language could also be found by persistence (longitudinally and transversely); however,
concepts of the concept language (and how these relate to the input language) are far less clear.\label{hubscales}}
\end{center}
\end{figure}

 Only at the level of hubs is there a plausible metric notion of
  distance or compositional similarity.  Although any expressed
  attribute can be compared in terms of the alphabet of its smallest
  fragments, order generally renders sequences unique, so there is
  little or no mutual information to go by.  Only by fractionating
  sequences into an alphabet of disordered parts can be measure
  similarity in a consistent way.  The collection of all attributes
  $\phi_n$, for each hub $H_i$, may possess mutual information in the
  alphabet of $\phi_n$ with respect to every other hub $H_j$.
  Clusters of sentence events, joined to context hubs, may therefore
  be measured as `similar', near or proximate to one another if they
  overlap in their support of $\phi_n$ express from below (see figure
  \ref{hubscales}).

\begin{table}[t]
\begin{center}
\begin{tabular}{|c|l|l|}
\hline
\sc Fractions & \sc Text & \sc Bio-informatic \\
\hline
$\phi_1$ & words & bases \\
\hline
$\phi_{2,3}$ & names & codons\\
\hline
$\phi_\text{short}$ & concepts & genes\\
\hline
$\phi_{n>3}$ & embellished concepts & peptides\\
\hline
$S_t$ & events & proteins\\
\hline
$\{\phi_n\}^{(+)}$ & context & mixture\\
\hline
overlap $\{\phi_n\}^{(\pm)}$ & themes & species\\
\hline
Narrative & narrative & bio-process\\
\hline
Directed regional & Meaning & Functional\\
stories  & and intent & adaptation\\
\hline
\end{tabular}
\bigskip
\caption{\small Approximate identification between text and bioinformatic representations of process
narrative. The similar scales arise likely for the same underlying reasons: scale separation of information
is critical to stable spacetime invariance.\label{table2}}
\end{center}
\end{table}

\subsection{Concepts and themes}

It's worth a brief digression to clarify some narrative terminology.
In literature one distinguishes the notion of concepts from
themes\cite{concepttheme}.  Since we are using data from natural
language narratives, this issue is important because it captures
a phenomenon of scaling.  Although the distinction is loose and based
on preferred interpretation in natural language, it turns out that
there is a natural way to distinguish these based on scaling.  

The difference between a concept and a theme lies in their semantics:
while a theme captures a broader area, a concept limits itself to a
narrow and particular idea (i.e. input pattern)\footnote{That is not
  to say that concepts could not take on the role of themes and vice
  versa, on different levels once they have been assigned proper
  names---because the proper names are concepts which then represent
  the themes.}. In a spacetime sense, they arise differently, as
patterns belonging to different processes.  Concepts are found in the
discrimination of {\em longitudinal} input language patterns (time), by
searching for invariants. Themes, on the other hand, are found from in
the {\em transverse} correlations between disordered contexts (space).

We need a term which refers to invariant characteristics
in the input---and `concepts' matches this well. Concepts have to be smaller than events in order for events
to be about concepts, so there a natural separation of scales.
\beq
\text{concepts } \phi_n < \text{events } S_t < \text{contexts } H_i < \text{themes } R_{ij}
\eeq
This view turns out to have a natural resolution based
on precisely the notion of spacetime process invariance---which is
easy to discover using the interferometry method.
In academic literature the concept of a concept is often associated
indistinguishably with `keywords'\cite{feldman1}. This fits with the
hierarchy implied from the process scales of the input stream.
Themes are associated with composite structures, based on mixtures of concepts representing
context of events. 

Now an interesting point arises: why are concepts defined in terms of
the input language and not the concept language? This can be
understood from the Spacetime Hypothesis itself---the most basic
concepts represent features of the spacetime processes around an
agent. Later, concepts might also come to represent phenomena involved
in processing and recycling interior representations of the world,
including imagined worlds, so the process can be repeated on a higher
level.  The bootstrapping of concepts, however, belongs to the
external sources.

We should always be careful to understand whether we are discussing
representations in the input language or in the concept language.
Elsewhere, the two languages may be identical, e.g. in studies of
text mining, but we distinguish them here with good reason: they
represent processes with their own invariants on different scales.  In
connection with this, the question which underpins the confusion
between the two is: how are proper names assigned to concepts in the
concept language, even as they are represented on a basic level in
terms of the input language?  If we play on the analogy with
bioinformatics, the question is like asking: how do we come up with
the name `Penicillin' for the concept of a particular mixture of
proteins and chemicals?

\section{Quantitative scaling analysis}

Based on the quantitative analysis in paper 1, we can now move on to
consider the discrimination of roles for different patterns, with the aim of
encoding sampled narrative in a geometry.  The problem becomes multi-dimensional due
to coarse graining and revisitation of concepts, and representing
simply scaling arguments becomes more of a challenge.  Using the same
data sources as in paper 1, we can nevertheless begin by looking at
how the formation of co-activation structures scales. This assessment
could have been extracted from the procedures described in paper 1 (no
new analysis is required), but they were not directly relevant there.

The Spacetime Hypothesis leads to a straightforward classification of
variables and separation of concerns, through the four semantic types. The goal of this account
is therefore limited to presenting the data through the lens of the model.

\subsection{The basic quantities}

It seems self-evident that the more data we have, the more potential
there is for extracting concepts. With that in mind, its sometimes
expedient to express derived measures relative to total input language
word count. This is not the measure of proper time (sentences) which
drives the narrative, rather it corresponds to the work done by the processing of the input language.
Elsewhere, its useful to measure relative to the memory samples, which related
to the proper time. Some measures are thus plotted relative to the number of hubs,
which is the relevant discriminator for the concept language.

The notation for the elements is summarized here:
\begin{itemize}
\item Word count $w$ for each narrative.

\item The number $|H|$ of hubs $H_i$ that are collated from each narrative,
for index $i$ running over the distinct hubs.

\item The number of interconnections between hubs $H_i$ which could range from
$0 \ldots |H|(|H|-1)/2$.

\item The number of sentences contained by hubs has no symbol.

\item The number of fragments expressed by sentences and hubs is simply written as a set $\{\phi_n\}$,
where $n$ is the number of words per fragment.

\item The proximity of hubs to one another is measured by the overlap (or interference), 
of hubs in different contexts or legs of narrative. This can be measured within the same
narrative or between different narratives.

\end{itemize}
As is typical in scaling theory, from the Buckingham-Pi theorem, there is a key dimensionless variable
that controls many aspects of the highly non-linear behaviour\cite{scale1,scale2}.

\begin{figure}[ht]
\begin{center}
\includegraphics[width=7.5cm]{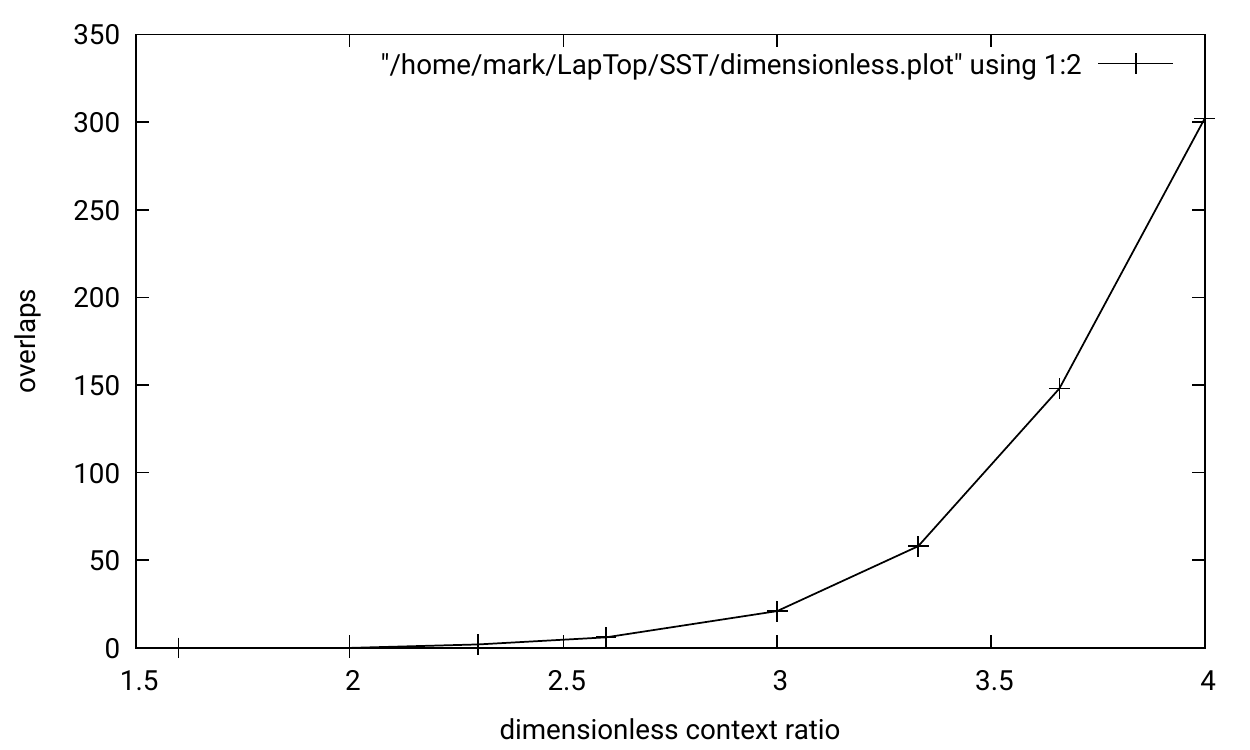}
\caption{\small The increase in overlap occurs around the transition $\nu\sim 2-3$, which
corresponds to the size of fragments which are repeated significantly
in different context characterizations or $\phi_n$ mixtures.\label{dimensionless}}
\end{center}
\end{figure}

\begin{definition}[The context ratio $\nu$]
  A dimensionless ratio, which characterizes the sampling and memory
  representation process, measured by comparing word counts for a
  typical sentence with a sum length of all fragments retained in
  fractionated form as its context (hub). The ratio of average skimmed fractions
  $\phi_n$, for all $n$, divided by the average length of sentences in
  the local narrative region: \beq \nu = \frac{\langle \sum
    \phi_n\rangle }{\langle S_t\rangle} \eeq
\end{definition}
This ratio of scales can vary throughout a narrative, as the lengths of sentences
varies, and so on. A more sophisticated sampling agent could adapt this ratio
to improve the efficiency of its cognition, in principle. To keep matters
simple, we don't try that here---but we can see the effect of varying the
ratio (see figure \ref{dimensionless}). Concerning the larger significance
of this ratio, we see that it is not a self-scaling (probabilistic) measure that
characterizes the important processes, but rather a comparison of different
dynamical scales measured from the basic characters of spacetime variation.
This point alone favours the Spacetime Hypothesis's deviation from a 
probabilistic approach to recognition.

Owing to the special importance of 2,3 fragments, a critical region for this.
A typical value of 2-3 turns out to predict when significant associations
can be made. Below this limit, association does not become a useful tool
for learning across narratives. For significantly larger values, spurious
associations form all the time, leading to a blurring of concepts and themes
into a `grey goo' (maximum entropy).

The more conventional measures of graph structures, typical for
example in percolation studies, are the node degrees $k_i$ for each
agent $A_i$. These can be separated with respect to spacetime types
1-4 for each node in the graph, which refer to different scales.  From
this node degree, one could estimate the level of percolation using
the theory of Newman et al\cite{newmanreview}. This does not turn out to
be illuminating. The graphs produced by the spacetime method are not without intentional structure,
so its unclear the extent to which they might be considered random graphs.

As in paper 1, serial text sources are scanned, sentence by sentence,
using a realtime process, with a continuous forget-rate. This ranks
sentences and fragments by a measure of importance based on spacetime
scales. In paper 1, the sentences were considered for their accuracy
in summarizing the intent of a narrative. Here, we take the fragments
and extract their implicit relationships, according to the spacetime
model.  By treating each fragment as a promise theoretic agent, and
attributing four basic promise types based on the spacetime geometry,
this builds an effective geometry from a graph of the promises made
between parts. Interferometry is used in two ways (along the narrative
and across different narratives) to search of stable invariants of the
narrative as a semantic process\cite{spacetime3,cognitive}.  Some
sample data sources are shown in table \ref{tab1} for illustration.
For the purpose of understanding the role of bias due to familiarity
with the narratives, some of the chosen texts were written by the
author, while others were merely `known of' and others were completely unknown.

\begin{table}[ht]
\begin{center}
\begin{tabular}{|c|c|c|l|}
\hline
Words & Name\\
\hline
5193 & Thinking in Promises 1\\
2925 & Thinking in Promises 2\\
4945 & Thinking in Promises 3\\
2897 & Thinking in Promises 4\\
5445 & Thinking in Promises 5\\
5455 & Thinking in Promises 6\\
10190 & Out of the Fog (novel)\\
112538 & The Promised Land (diary)\\
125932 & History of Bede\\
192106 & The Origin of Species (6th)\\
208458 & Moby Dick (Novel)\\
216842 & Smart Spacetime\\
261132 & Slogans (Novel)\\
\hline
\end{tabular}
\bigskip

\caption{\small A few of the sample texts. The most coherent behaviour
  is observed in the book Thinking in Promises, which concerns a
  narrow specific subject, like a typical text book. The other books
  are more expansive in their topics. Novels are the most expansive.
  Note there are minor differences to the counts shown in paper 1, due
  to the minor editing out of copyright informations from certain
  texts to eliminate noise.\label{tab1}}
\end{center}
\end{table}

\subsection{Sparse graphs enable separability}

We begin by performing some basic measurements of the data,
to get a feel for the important scales and possibly trivial
quantitative relationships that can be used to make sense of the more
complex results later.

Story summarization, or extraction of a chain of events into a
sequence of nodes forming a trajectory, is the basic process by which
narrative is ingested by the system.  Apart from the natural and
approximately linear relationship between story trajectories and word
count (sample size), one would not expect whatever concepts emerge to
follow any obvious pattern, on the basis of a purely quantitative
measure like word count---any more than one might expect the shape of
someone's nose to be related to their overall mass.  Concepts are
signature features of a narrative, and are, by definition, not likely
to be regular statistical phenomena.  Some patterns might nevertheless
still fall into a few general classes. The distinction between factual
texts and fictional ones stands out here.

Sparseness is what allows separation of concerns to be effectively
maintained. Should a cognitive
agent ever manage to saturate its memory representation, it would
spell doom for its reasoning capabilities. Sparse connections enable
close to linear growth of linkage within the quadratic space of possibility.

\begin{figure}[ht]
\begin{center}
\includegraphics[width=7.5cm]{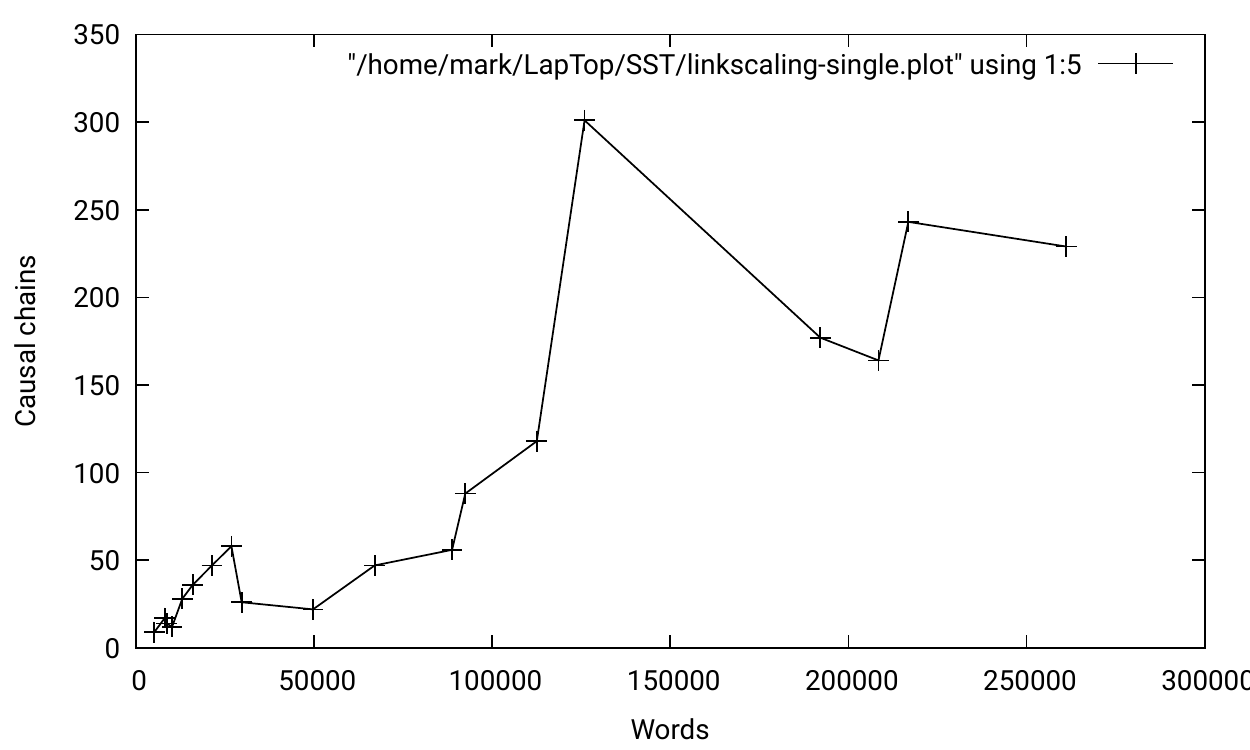}
\caption{\small The number of causal precedence/antecedence links
  between events retained by scanning, as in paper 1, indicating the
  length of story trajectories.  One might expect this graph to be
  approximately linear, given narratives with constant `significance
  density' throughout. The anomalous point in the middle of the graph
  (from the History of Bede) suggests a document with a lot of
  repetition of concepts. Note: the joining line is for ease of
  reading and does not imply interpolation.\label{follow}}
\end{center}
\end{figure}

Figure \ref{follow} shows how the length of story sequences (connected
sentences) grows with input size.  One expects this graph to be
approximately linear, based on an assumption of constant average
`significance density' throughout each narrative, however it's not quite
linear due to the variable sampling rate. The anomalous point
in the middle of the graph (from the History of Bede) suggests a
document with a lot of repetition of concepts.

\begin{figure}[ht]
\begin{center}
\includegraphics[width=7.5cm]{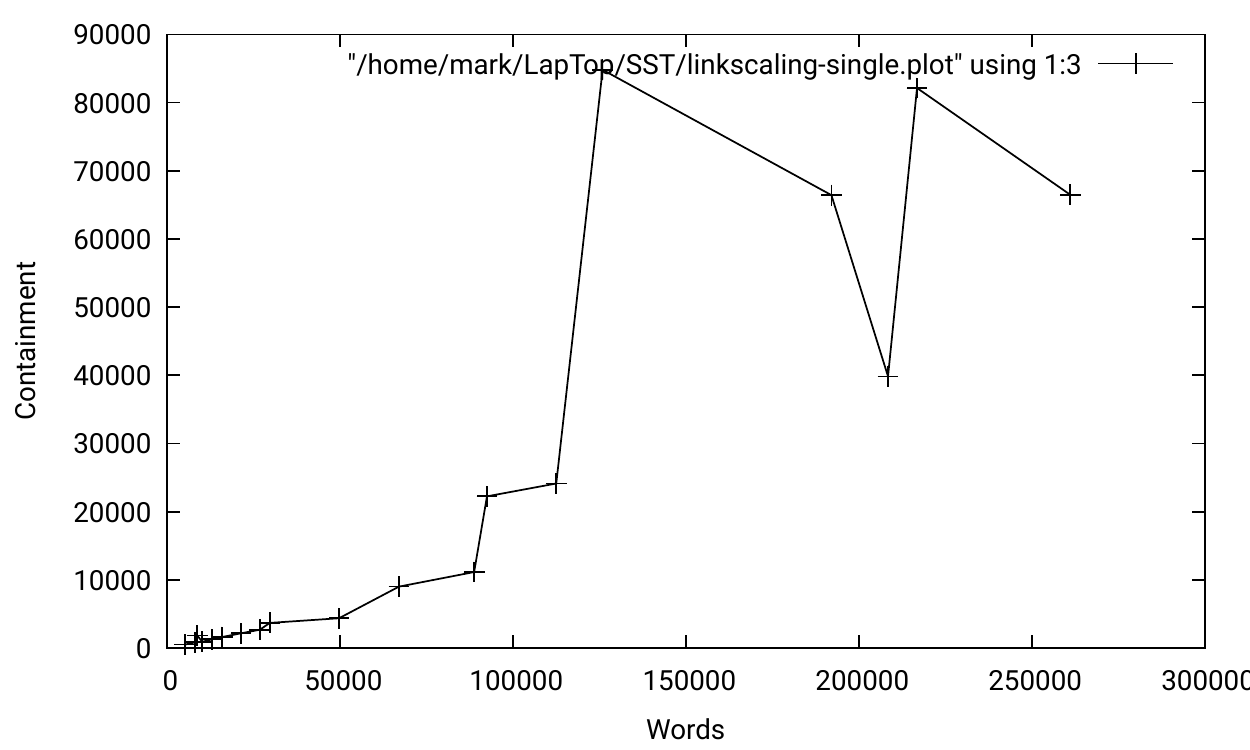}
\caption{\small The level of containment by context hubs for events and short fragments $n \le 3$
  retained by scanning, as in paper 1.  by the hierarchical
  construction and constant sampling rate of fragments aggregated into
  hubs. Note: the joining line is for ease of reading and does not
  imply interpolation.\label{contain}}
\end{center}
\end{figure}

Figure \ref{contain} shows the number of meaningful events and short
fragments `contained' by narratives, pruned by importance above basic
threshold (equal for all), for a particular $\nu$.  The exact number
is dependent on many factors, which were varied over the trials, but
the scaling pattern is similar in each case.  The arbitrary scale $\nu$
which makes this choice acts as an independent variable.
It's role is to limit noise from spurious words, and
I'll comment on this further below in section \ref{stability}.  The
pattern shows that there is no obvious connection between scale and
fragment density.  The beginning of the graph comes from the
chapter-by-chapter analysis of a single narrative and shows that the
fragment growth is just sublinear. As the other longer narratives are
added there is a sharp fall and a rise again. This could be because
certain texts contain a lot of repetition of terms. On anomalous case
is the History of Bede, which is a litany of proper names and events,
which therefore seems to have a lower diversity of stable fragments
per word length than other texts.

\begin{figure}[ht]
\begin{center}
\includegraphics[width=7.5cm]{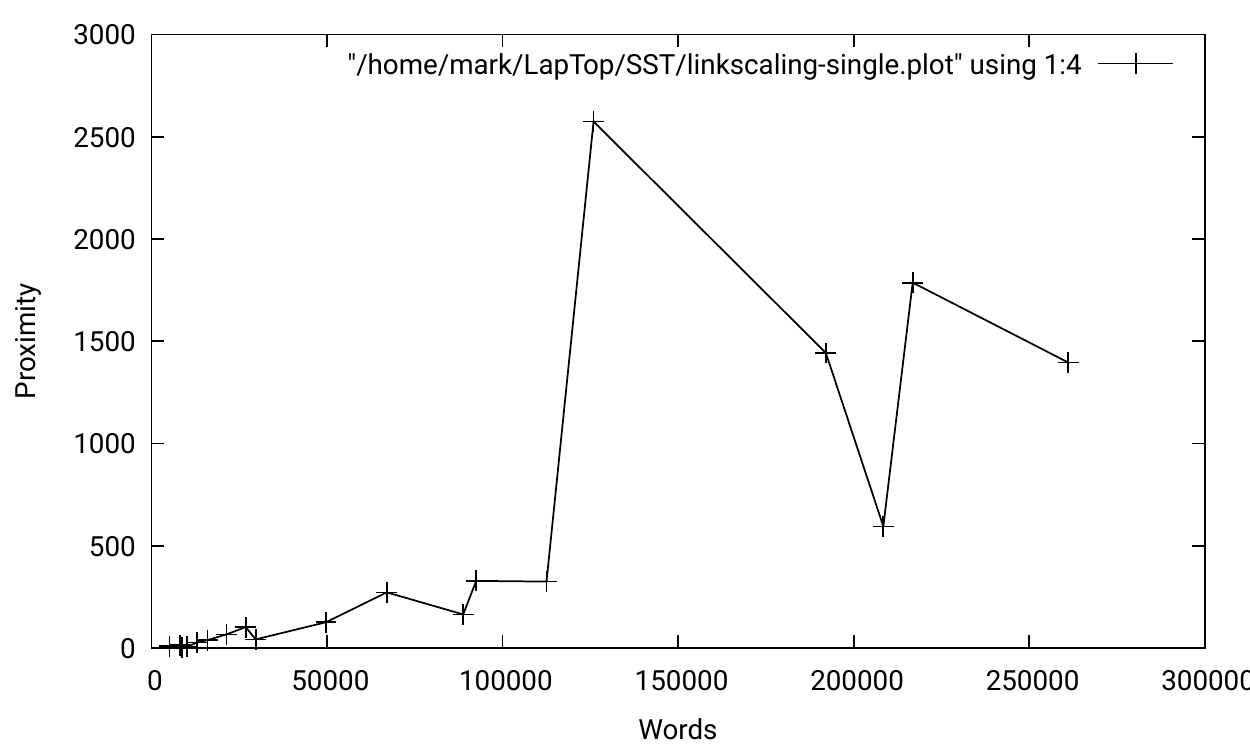}
\caption{\small The number of proximity links added between hubs
  during `sleep' post processing indicating the degree of concept
  formation by overlap. It seems hard to predict the extend to which
  graph might scale with words, as the overlap between contexts
  depends on so many causal factors that are absorbed by coarse
  graining. Note: the joining line is for ease of reading and does not
  imply interpolation. The anomalous point is once again for Bede, a
  narrative with a lot of repeated fragments.\label{near}}
\end{center}
\end{figure}

\begin{figure}[ht]
\begin{center}
\includegraphics[width=7.5cm]{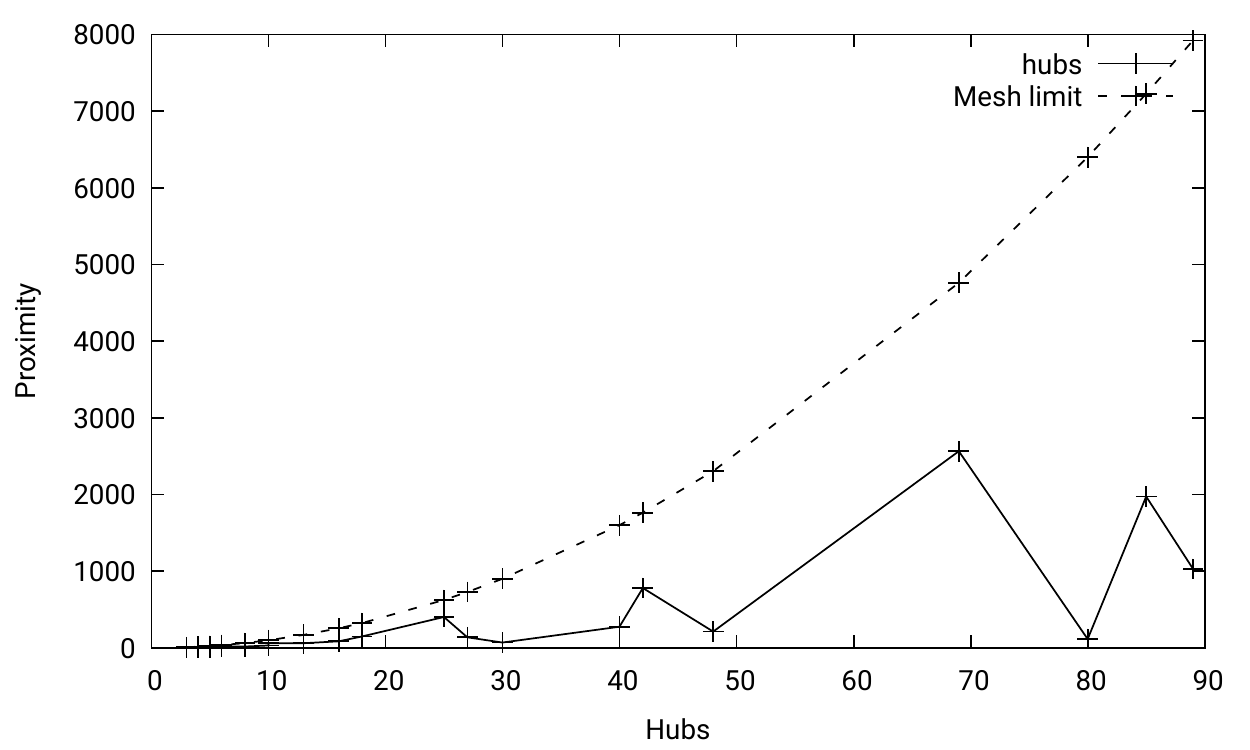}
\caption{\small The proximity graph density versus its maximum
  $N(N-1)$ shows that as the inputs grows, the connections remain
  sparse for effective separation. As the limit is approached, theme
  separation dissolves into a maximum entropy state. The data points
  are well under the saturation level, so the graph remains
  sparse.\label{sparse_near}}
\end{center}
\end{figure}

More interesting is the graph of coincidental overlaps, implying
proximity of hubs to one another (see figures \ref{near} and
\ref{sparse_near}).  It's not obvious a priori how such a graph might
behave. First of all, one has to accumulate stable fragments, then
they have to be repeated in similar patterns in order to end up with
similar hub contexts. Then there is the unpredictability in combined
importance of the fragments leading to the keeping of sentence events.
All of these factors come together in this figure.

One effect that seems to be apparent from the experiments (which is
evident though not conclusively in the data) is that as a greater
amount of unfocused text is added, concepts come and go. Regions of
related context emerge but are later merged with others.  Without a
way of keeping regions separated, the graph of contexts and knowledge
will eventually percolate in all directions, leading to one giant
concept cluster.  This corresponds to the maximum entropy state for
the effective alphabet of contexts.  As observed in paper 1, too much
information is therefore as bad (if not worse) than too little.  Too
little information may be survivable, but too much may be
unrecoverable\footnote{In science fiction stories it has often been said
  that humans use only 10\% of our brain capacity. Imagine what could
  happen if we could use all of it. The answer seems clear. If we were
  able to somehow fully utilize all of the nodes in a brain, this
  would be a state of maximum entropy and we would cease to function
  long before we reached that level. Sparse utilization is not a bug,
  but a feature of the implementation method.}.

\subsection{Region formation by hub interferometry}

Hubs represent contexts. Interferometric overlaps between
contexts can therefore lead to proto-concepts or themes. They emerge
from single sources of narrative. Hubs are too specific to allow
concepts to emerge that are independent of context. Regions are what
one might assume to correspond to more evolved concepts over time.

Context is associated directly with the scale of the mixture of fragments $\phi_n$.
Concepts are the spectra condensed from the soup of concepts, by virtue of an
interaction chemistry, which is implicit in the spacetime properties
of the input stream. The allusion to molecular chemistry is no accident.
The same principle applies, whether the language of fragments is
chemical, phonetic, or lexical: symbols are symbols.

The natural place for concepts can be understood at the scale at which
one can observe similar processes (event sequences) derived from
partially-similar invariant constituents, i.e. similar functions that span a set of
constituents. Partial events may be invariant concepts, so concepts must be smaller
than events.
Entire events will rarely become concepts, by a method based purely on
aggregation. The ability to form concepts comes from recombination of
the most elemental fragments, just as one sees in chemistry and
genetics.  If we take that lesson from here and apply it to chemistry
and biology, then it says that concepts correspond to invariant
molecular fragments (e.g. genes), while themes amount to the scales of
protein bindings and above, rather than codons, genes, or polymers
(figure \ref{coactivation2}).

So words from the input data stream do not lead to concepts by
themselves. Reconstituted fragments do.  How we end up giving names to
concepts is an entirely different discussion that belongs to the scale
of the emergent commentary language.  The association of proper names
to concepts is a separate process, as is the association of a shape or
image to a concept.  Examples of concepts (occurrences in topic map
parlance\cite{topicmaps,TAO}) express exemplifying events which have not been broken up
and reconstituted---but rather remembered as raw data (`as is').

\subsection{Post processing of narrative graphs---lateral thinking}\label{sleep}

Hubs link together collection samples of fragments associated with a
sample point of sensory context. Hubs form effective basis elements of
narrative, in the sense of a matroid pattern.  The sum of the $\phi_n$
components also acts as a proper name, expressed in the input
language. A different alias could later be given in the concept
language too; however, that might be premature. The hubs don't usually
represent invariants---only parts of them are invariant.

As we aggregate fragments (some of which might be concepts) into
repeated clusters, we can find transverse invariants by looking at the
overlap regions of these mixtures. This can only be done once hubs
have all formed, and is best performed when the system is in rest,
i.e.  when it is not accumulating new context. So during this `sleep
phase' it makes sense to run through all recent memories and compare
them to older ones that are `coactive' with respect to
context\footnote{It's interesting to speculate whether this separation
  of scale might be a reason for dreaming in animals.}.

By stimulating the component fragments that represent context, at the
bottom of the memory hierarchy, overlapping contexts can be
identified, measured for their overlap, and joined together with a
weight representing how much overlap there is. These joined up hubs
are then `near' one another and form weakly linked regions whose
conceptual mixtures are similar of proximate to one another.

Regions are the only structure (apart from individual concepts) that
can span different narratives.  The impact of these cross connections
is therefore profound. Repetition under changing context implies that
overlap fragments are invariants. As collections of invariant
concepts, they therefore act as `themes'.  Moreover, since we
associate stories with pathways or trajectories through fields of
concepts, and hubs with contextual environment for key events, one can
expect events to be weakly linked through hubs that are similar too.
One event can trigger the idea of another by jumping from one context
to a similar one, and looking for compatible events in this way. This
is `lateral thinking'.

Stories---or artificially generated episodes---begin as separate silos
of knowledge, unrelated by different experiences.  How do they become
related?  The sum all all reasoning pathways, starting from any given
event or concept, is combinatoric in nature and grows exponentially
and discontinuously with each selection of a link in the chain.

It's unclear, at this stage, at what scale such a story might be told.
However, if we assume that simple-minded stories (in a kind of pidgin
language) might result from the recombination of stable events from
the original input, they we can search those possible pathways within
the spacetime hypothesis to generate new stories from the vantage
point of a godlike observer.  They might not be in the eventual
commentary language of the cognitive system, but we can't wait for
that language to emerge to finish this paper, so the compromise will
have to suffice.

In other words, instead of rushing to interpret text using concepts of
the input language, instead we should think of concepts as accumulating
somewhat like molecular genetic structures: small components have
similar functions on a low level (see table \ref{table2}). Their combination (and
recombination) can lead to other expressions, and these overlap with
one another through the language of fragments (e.g. genes or codons).

Overlap in the chemistry of hubs can be `cached' graphically to assist
in the computer model, by forming semi-permanent weighted links
between them. The links have spacetime type `NEAR', and serve to
measure the proximity by similarity of hubs from one another.

A measure of distance can only apply to multifragment hubs, because
similarity is only meaningful where there are different sets to count
and compare to one another. There's no naturally meaningful way to
measure the distance between sentence events or fragments, which are
atomic symbols. So the sub network of type NEAR forms clusters that
aggregated into undirected globules of associated co-activation
contexts. On the basis that similar context implies similar
interpretation, the spacetime hypothesis then basically says that new
concepts would form around these clusters on a new scale---that could
be identified as the concept language: whose vocabulary is an
accumulation of micro-concepts.

Comparisons of hubs, i.e. contextual admixtures, can be performed across
narratives as well as along them (both transversely and longitudinally).
The imagination hypothesis suggests that we could tell new stories
in this way---by jumping contexts to fill in a chain of reasoning
with either sequential concepts or event playback.

\subsection{Emergent process scales---a natural relevance horizon}\label{scales}

The space of control variables in even this simple model is large.
The dimensionless context ratio $\nu$ plays a role in the possible
size of overlap regions between hubs, because it contains the amount
of short term memory available to cache recent context.

Initially, a self-scaling of interactions was used to compare all hubs
on a compressed scale---typical of probabilistic methods and self-similarity studies.
When the numbers were scaled using relative to self, i.e. as a fraction of total
sample, the result was highly irregular, because the sample sizes themselves
had such a varying absolute size, as measured in units of the pattern alphabet (words).
This led to highly unstable results.
Empirically, looking at the numbers, we find precisely these scales represents in
clearly separable terms.  There are two horizons: nearest neighbours,
which seem to reliably correspond to:

\bigskip
\begin{tabular}{|l|c|c|}
\hline
Random     & $< 1\%$        & weak (over horizon)\\
Meaningful & $1-10\%$       & local (relevant)\\
Repetition &  $\simeq 50\%$ & self (ignorable)\\
\hline
\end{tabular}
\bigskip

Some success in matching regions could be obtained, but only by
artificially introducing an event horizon for random overlaps, which
felt unsatisfactory.  In the small regime, the overlap distances fell
into principally three scales: larger co-activations (around 50\%),
presumably from persistence of a single event in short term memory,
being sampled twice and leading to artificial duplication.  Then there
will always be random overlaps from small numbers of fragments that
just so happen to share parts of the same chemistry---this is spurious
and a result a combinatoric nature of patterns in the input. The
final, more interesting, kind is due to a significant correlation in
the occurrences. Expressing a raw statistical basis for this
correlation is not simple, owing to levels of obfuscation through
importance functions, threshold selections, and subsequent
aggregations.  Pruning the category 1 weakest links, which are
essentially by random chance, the integrity of regions is more robust.
The effect of the strongest links can be essentially neglected. This
usually only happens when two events occur so closely together that
they share the same context. There is an effective uncertainty relation
here between the overlap $\Delta H_i$ and the context sample:
\beq
\Delta H_i \Delta t_j \gtrapprox \nu\, \delta_{ij},
\eeq
using the Kronecker delta of the proper-time sample points,
which arises from the fact that locations represent coarse grained
(non-local) regions over an aggregate scale\footnote{The change in
hub constituency over a proper interval is somewhat analogous to a 
canonical momentum in mechanics.}.

On more careful consideration, an appropriate solution was to return to a proper dimensional
analysis of the intrinsic scales, and introduce the dimensionless
context ratio, as the relative scaling
multiplier instead of trying to eliminate scales altogether.  During
the experimentation, three regions in particular were examined.  A
parsimonious region of small buffer size $v < 1$, where there could be
little overlap, a region of $v\simeq 3$, and a region of large overlap
$v > 5$, which behaved quite differently. 
The critical value of around
$v\simeq 3$ presumably arises from the role of fragments of length 3
in generating meaningful overlap (see figure \ref{dimensionless}).
Fragments of length 1 carry no ordering context. Fragments of length 2
carry a little, but the optimal length is 3\cite{cognitive4}, and
longer fragments almost never repeat.

\subsection{The scaling of hub overlaps during interferometry}

The significance of overlap is to find those fragments which are
non-unique and which therefore represent invariants, and therefore be
considered concepts alongside their longitudinal compatriots.

The suggests a more natural interpretation of the components of the input language:
\begin{itemize}
\item The short fragments, or spectral contents of mixtures, which
  participate in overlap represent proto-concepts.  Each of these
  corresponds to a potential symbol in the concept language. Over many
  learning episodes, one might imagine these become stable.

\item Longer fragments and non-overlapping remnants can be quickly
  forgotten as past context.

\item The repeated admixtures of short fragments can be associated with {\em themes}.
Themes are thus mixtures of concepts that convey broader {\em intent}.
Higher level `intentionality' (as we understand the high level
concept) emerges through repeated themes.
\end{itemize}
The associative distance between the input and the concept languages is thus remarkably
short. However the input may be stored, its short fragments become new symbolic
invariants---like sieving the input for gold nuggets. Those nuggets must eventually
form higher representations from which concept language could emerge\footnote{Note, one
shouldn't assume that the concept language is written or oral, it could be entirely visual,
and need not even be communicable between cognitive agents.}.

\begin{figure}[ht]
\begin{center}
\includegraphics[width=7.5cm]{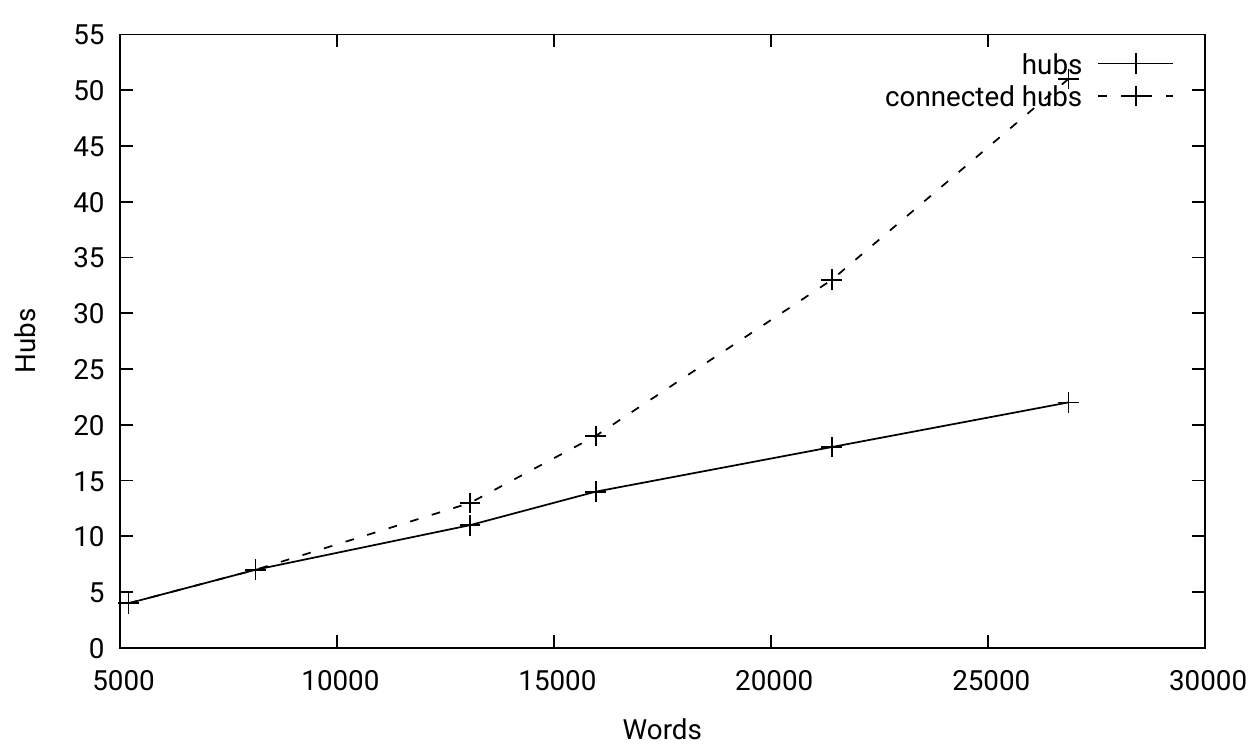}
\caption{\small The number of distinct hubs that link events to clusters of $n$-phrases,
and the number of those that overlap taking samples from the same narrative in increasing
step sizes. The solid line connects the number of independent hubs, while the dashed lines shows the
number of connected hubs, which could rise to the square of the number.
The connecting lines are for ease of visualization and do not imply interpolation,
since the values change discontinuously.\label{hubscaling2}}
\end{center}
\end{figure}

Figure \ref{hubscaling2} shows how the number of coherent clusters
(hubs) from one example trial grows on successively sampling larger
amounts of the total narrative. One sees a linear growth and a more
quadratic shape. In a sparse process, these can be clearly identified within a
potentially quadratic process of cluster overlap. Taken from a single
narrative, based on a text book (Thinking in Promises) with a clear
subject matter, it's gratifying to see this level of predictability.
However, we shouldn't get too excited: this all falls apart once
different kinds of narrative and different lengths of narrative are
concerned.  The linkage is slightly superlinear, indicating
non-trivial connectivity, inherent in creative recombination
processes\footnote{Superlinear scaling has been associated with
  recombination attributed to innovation in studies of cities, for
  example \cite{bettencourt1,bettencourt2}.}.

\begin{figure}[ht]
\begin{center}
\includegraphics[width=7.5cm]{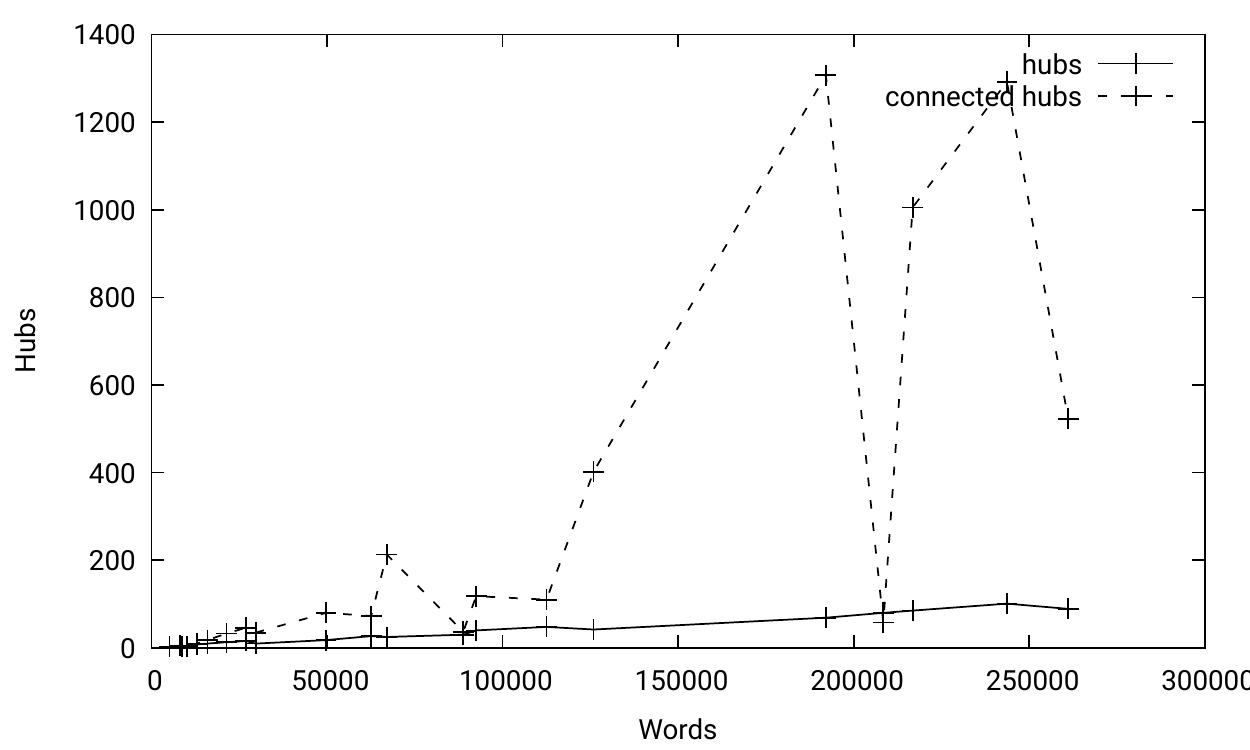}
\caption{\small Compare to figure \ref{hubscaling2}, now adding several different narrative sources, of longer
length and differing types. The neat scaling relationship in figure \ref{hubscaling2} is all but eradicated,
indicating that it was probably a special case rather than the general rule. One would not expect
regular scaling for a percolation in a random graph.\label{hubscaling1}}
\end{center}
\end{figure}

\begin{figure}[ht]
\begin{center}
\includegraphics[width=4cm]{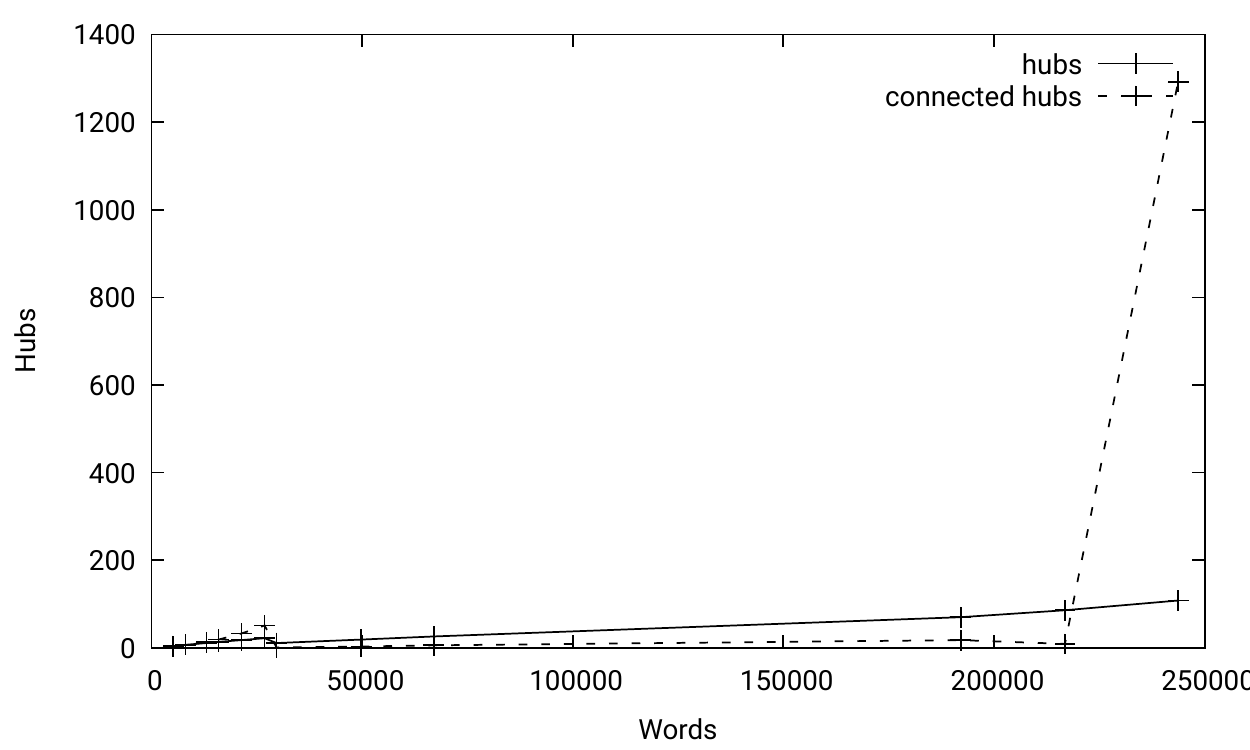}
\includegraphics[width=4cm]{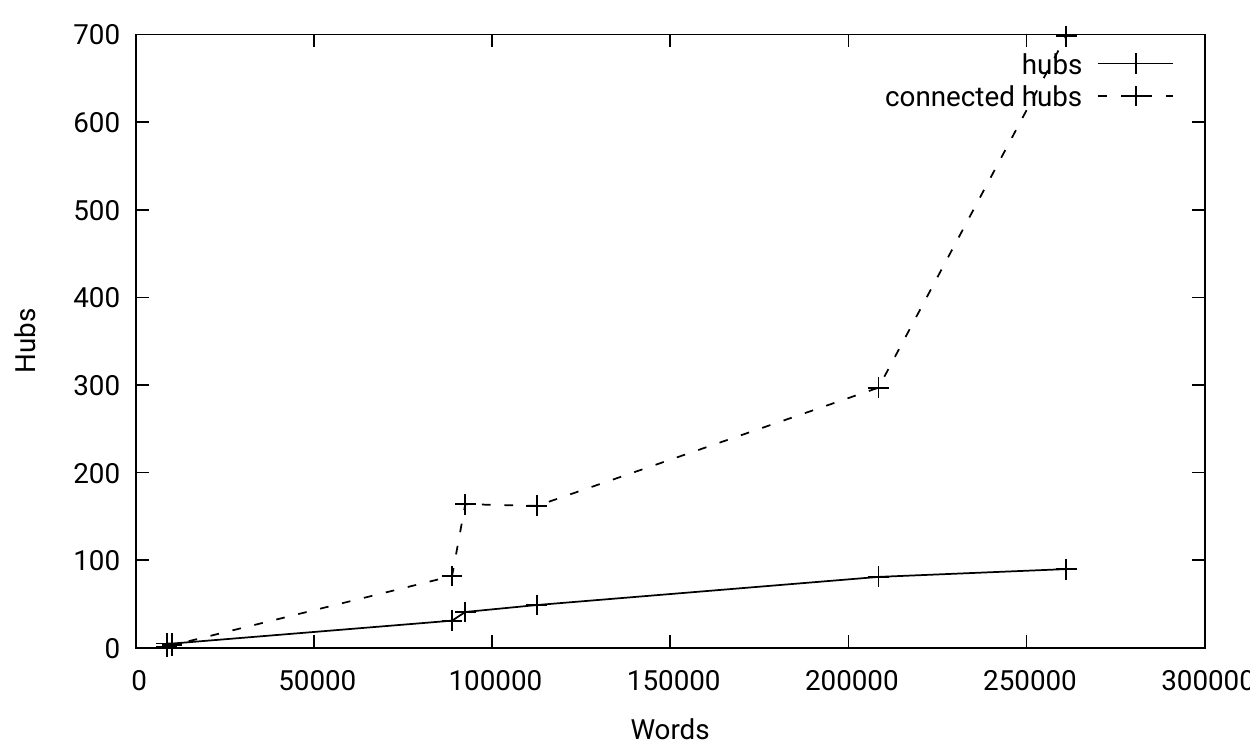}
\caption{\small For convenience, the hub scaling data in figure
  \ref{hubscaling1} separated by textbooks (left) and by fiction
  (right). No convincing quantitative pattern reveals itself for these
  two categories: the principal differences lies in the semantic
  chemistry of the components not in their quantitative measures.\label{hubscalingsep}}
\end{center}
\end{figure}

\begin{figure}[ht]
\begin{center}
\includegraphics[width=7.5cm]{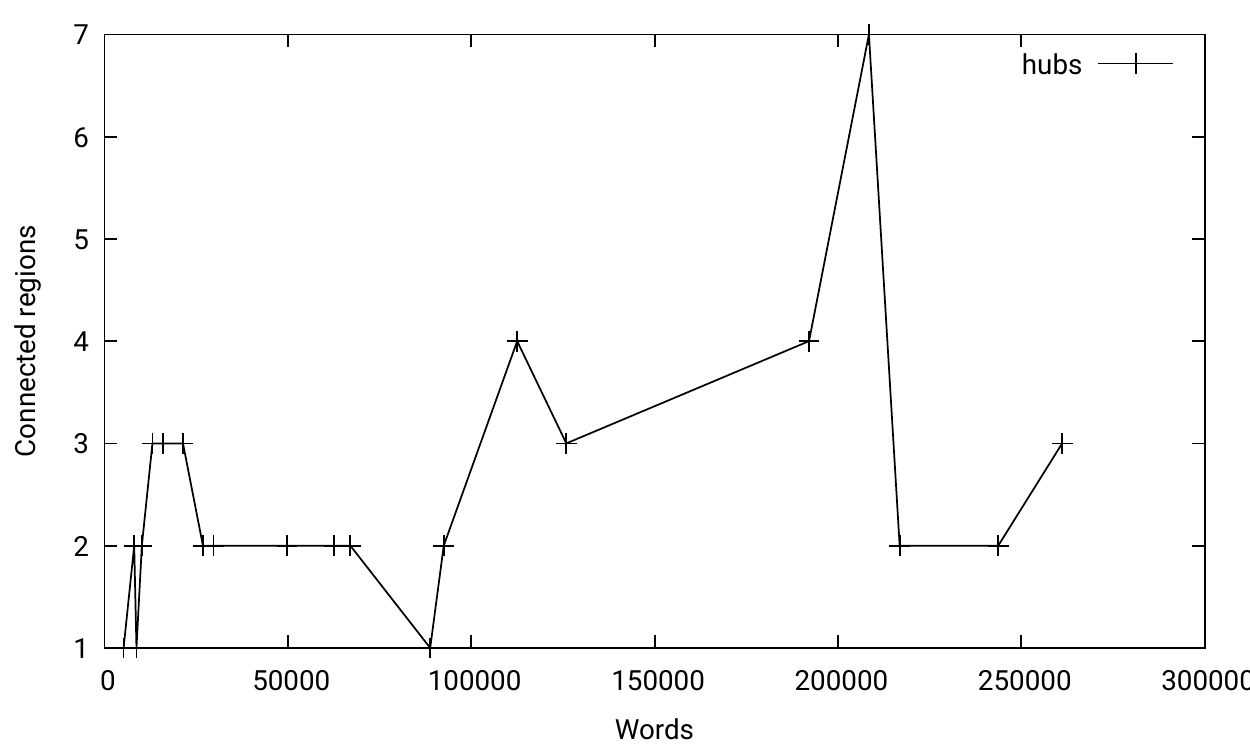}
\caption{\small The number of regions (connected hub clusters that are
  deemed proximate to one another) versus the number of words sampled.
  One might expect this to conceal more noise, since stragglers could
  be absorbed into larger groups. However, from this small number of 20 samples of up to 200,000 words,
this does not seem to be the case.\label{regions}}
\end{center}
\end{figure}

Figure \ref{hubscaling1} shows what happens to the same data once expanded with other examples. The neat
polynomial behaviour is just a corner of a bigger picture with spurious (and catastrophic) changes. And beyond
that, once the protected hubs are exposed to one another for overlap connections (figure \ref{regions}), 
The reason for this can be seen from the nature of the fragments in each case (see section \ref{regionsec}).

\begin{figure}[ht]
\begin{center}
\includegraphics[width=4cm]{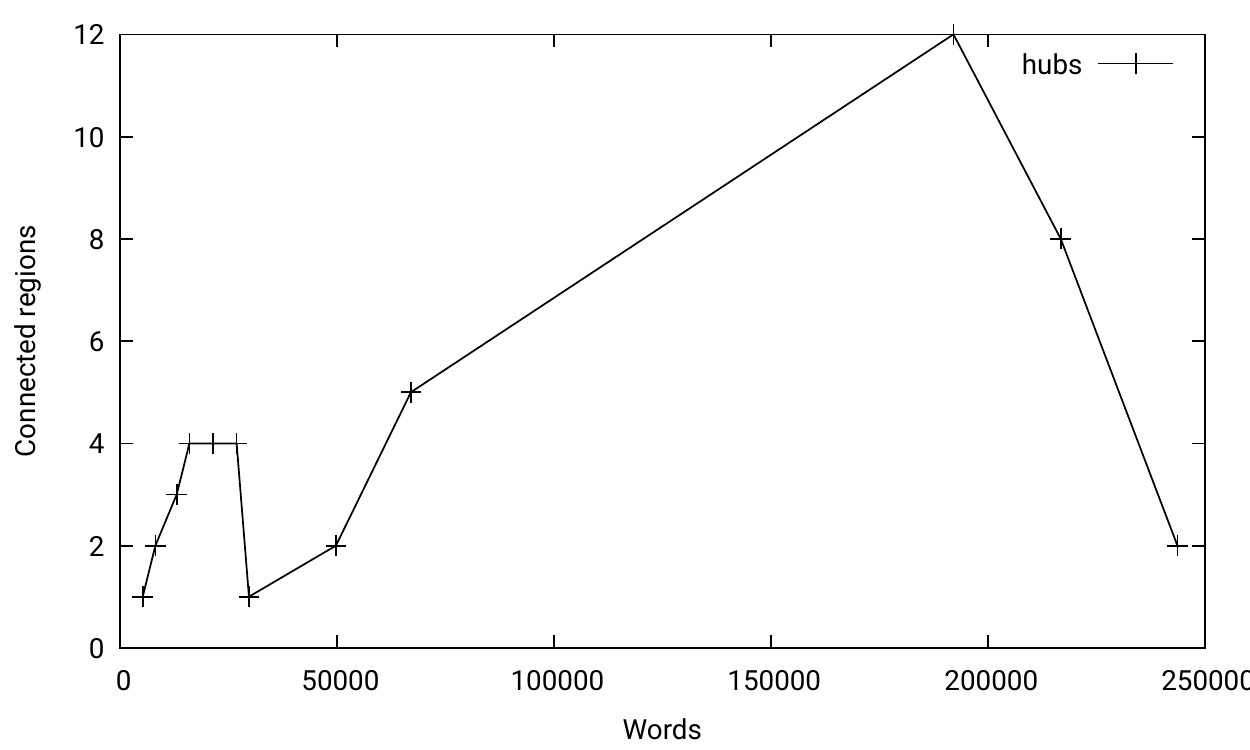}
\includegraphics[width=4cm]{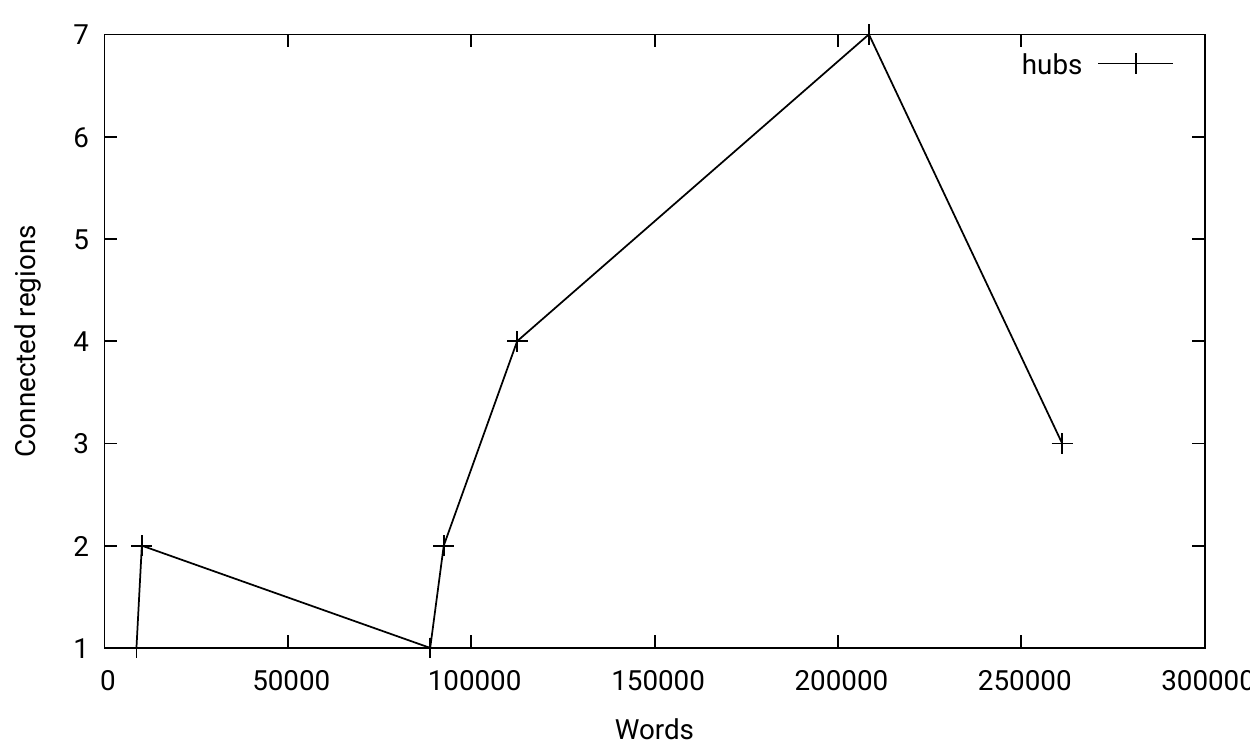}
\caption{\small For convenience, the region data in figure \ref{regions} separated by textbooks (left) and by fiction (right). The separation of classes doesn't add any convincing weight to the notion of fiction and fact being clearly separable, but the difference lies in the semantics, not in the quantitative measures.\label{regionssep}}
\end{center}
\end{figure}

Taking the example of the Thinking in Promises text, on adding a
significant length of text on a related (partially overlapping topic),
one might have expected the number of hubs to grow to extend the
number of concepts in a neat classification of knowledge. In fact, the
5 hubs collapsed into just two, throwing the whole story into
confusion.

After another experiment combining the proto-concepts of one with
another, 98\% of the connections were in the random category and
only 1.3\% in the range of plausible overlap. This suggests that this
might not be the mechanism by which learning representations grow.
Some mechanism to pin learning to a scale rather than growing out of
bounds might be necessary, for instance.  The role of specialized
scales (rather than scale-free behaviour) seems to around every
corner.

We see that there is a need for some process to
protect knowledge once formed.  The criteria for linking hubs must be
more subtle at scale, to avoid simply generating entropy.
Introducing a horizon to eliminate the random overlaps helps to stifle
the collapse---so this was taken to be standard practice thereafter,
and all further results are based on shielding hubs from small spurious
overlap (long range correlations).

\section{Interferometric extraction and stabilization of concepts}\label{stability}

The method of comparing a process to itself or to other processes with
small phase shifts is called interferometry. It's worth dwelling on
some of the technical details in arriving at the broad conclusions
referred to above, as these are non-trivial.

When parallel processes line up and produce the same symbolic outcome,
the addition of their outputs (called superposition) emphasizes the
result. When they fail to agree, the symbols accumulate more slowly
and gradually become demoted in relative importance. This approach can
be used in both spacelike and timelike directions to compare processes
based on ordered sensory streams of symbols.  Given a sampling process
in spacetime, one can choose to establish cumulative statistics either
longitudinally or along a timelike vector (corresponding to a Bayesian
update procedure), or transversely along spacelike vectors
(corresponding to a frequentist update procedure).  Both of these turn
out to have an important function in the spacetime method.

Along ordered sequences, longitudinal persistence can be used to pick
out fragments that are more invariant than random chance---intuition
suggests that these might correspond to `subjects' or `objects' in the
abstract sense of concepts, as one would likely base such
conceptualization on persistent phenomena.  These can be applied to
the fractionated samples acquired on each sentence arrival, taken as a
unit of proper time (see paper 1). The result is something like a spectrum
of concepts.

In disordered mixtures, transverse interference of the mixed
components can be used to pick out the chemistry of fragments, which overlap with others
due to independent similarity of pattern, from different sources.
Such overlap is even more significant, as it can cross over between
different narratives---but can only occur at the level of hubs, as
only these contain finite collections of symbols.  Thus, we have two
processes for the stabilization of key fragments $\phi_n$ and
collections of them $H_i$ respectively. Let's examine these in turn.

Taking a number of texts, and extracting the stable fragments leads to
a surprisingly cogent summary of what a text is about, but there's a
problem with this simple method: it assumes a knowledge of language.
The fragments have no meaning to a dumb sensor, only human observers
watching over the fragments can ascertain their relevance, given a
knowledge of the narrative content and the language in which it's
written.

\subsection{Longitudinal stability of fragments}\label{long}

A symbolic generalization of the method of wave
interferometry\cite{hecht1,mandelwolf,jenkinswhite} features in the
study, both here and in paper 1, as a way of separating signal from
noise.  The use of interference to highlight key $n$-phrase fragments
indicates that there must be an effective separation of scales between
fast and slow variables in a stream.  Looking for the slower variables
is a way to extract quasi-static `invariants' of the process.

\begin{hype}[Longitudinal invariants]
Longitudinally persistent $n$-phrases $\phi_n$ may correspond to important
subjects in a data stream.
\end{hype}
Indeed, the somewhat unexpected conclusion of the multiscale analysis
is that concepts have their origins in fact short fragments of the
input language---regardless of how they might be represented and play out in the
final concept language.

Consider a few examples, from the data, to illustrate this.
In the {\em Thinking in Promises} book, the set of terms
that reinforce by repeated use include:
\begin{quote}
\small
make, theory, promise(s),agent(s), this, information, world,
delivery,continuous, service(s)
\end{quote}
This is perhaps a surprisingly small number of words condensed from a
rather long text, but it is the principal remainder of longitudinal interference.
Comparing these to a human `cheat' knowledge of the book's contents,
the less common words (promise, agent, information, delivery,
continuous, service) do indeed correspond to key elements of the
book's subject matter, The individual subjects of the narrative (which
are players in the story it tells) would be like the {\em dramatis
  personae} of a play.

A second example from the longer book {\em Smart Spacetime}, on the other hand, yields a much longer set of terms.
Quoting only a few here:
\begin{quote}
\small
energy  spacetime  basic  scales  few same  interest  story  motion  semantic  add  way  conservation  deal  truth  processes so-called quantum understanding  theory computing information changes universe memory physics intelligence  cognitive reasoning explanations logic ...
\end{quote}
Apart from some spurious yet common padding words, these are all pertinent
`personae' within the text. So, while the longitudinal method does pull out good candidates,
it doesn't eliminate all noise.
From a sample of 269 invariants, there were trivial fragments and central fragments,
simply by persistence, with no other criteria in play. Looking on with the benefit
of evolutionary language skills, we see that persistence correlates surprisingly well
with central players in the text, offering another indication as to how certain
patterns become learned and associated with meaning.
\begin{quote}
\small
paths (85)
analogous (47)
using (121)
different (624)
life (36)
world (273)
conservation (39)
level (180)
 models (31)
 explain (69)
 much (141)
 artificial (59)
 behaviours (55)
 mechanics (149)
 mathematical (61)
 semantic (162)
 absolute (48)
 smart (85)
 scaling (142)
 story (131)
 pathways (43)
 approach (42)
 fundamental (70)
 computing (102)
 situation (23)
 spacetime (778)
 science (137)
 universe (139)
 semantics (124)
 relativity (128)
 quantum (267)
 cognitive (125)
 systems (216)
 intelligent (27)
 motion (335)
 theory (298)
 logic (36)
 function (35)
 thinking (72)
 intelligence (58)
 trajectories (21)
 thinking about (18)
 the existence (22)
 the dataverse (193)
 in some (22)
 the sense (27)
 process that (59)
 quantum mechanics (89)
 in a computer (19)
 at the same (49)
 the other hand (29)
 space and time (186)
 the same time (32)
 in the sense (20)
 in order to (89)
 at the same time (30)
 on the other hand (29)
\end{quote}
To the casual viewer, these innocuous choice might not stand out. However, on reflection the
list does indeed contain many of the central themes boiled down to simple phrases.
Using our godlike perspective, we see that these fragments are clearly concepts but not themes.
This shores up the thesis (see paper 1) that meaning arises by persistence against a background of noise.

One book which turned out to be an outlier in several of the measures
is {\em Bede's Ecclesiastical History of England}, which contains many
documented events and proper names.  Of 224 invariants found by
longitudinal interferometry, some were of a general nature, like:
\begin{quote}
\small
 both (105)
 little (39)
 night (66)
 knowledge (44)
 afterwards (70)
 departed (52)
 together (55)
 ecclesiastical (40)
 book (80)
 moon (48)
 christ (193)
 pope (106)
 died (100)
 written (40)
 history (63)
\end{quote}
Amongst the longer and more significant phrases of higher $n$:
\begin{quote}
\small
 the blessed (110)
 west saxons (26)
 the church (232)
 the mercians (51)
 the royal (20)
 king of kent (16)
 to the end (51)
 the english nation (40)
 came to pass (17)
 the apostolic see (17)
 the man of god (24)
 king of the northumbrian (16)
 king of the mercians (15)
 in the year of our lord (42)
\end{quote}
The method does indeed pick out key suspects in the narrative.

So, from the simple narrative scanning, based on spacetime patterns,
some important and dominant signals that stand out.  Persistence
appears to be quite a good selector of stable `conceptual fragments',
i.e. fragments that will eventually contribute to the stable regions
that are concepts.  This indicates that the spacetime interferometry
of fractionated language is a valid approach to a partially ordered
process such as languages.

We need to be clear about what this means. No one would expect such a
primitive algorithm to be able to extract anything like the nuance
that a human reader could---that would surely involve a deep mesh of
associative knowledge; yet the fact that such a simple idea can pick
out sensible core ideas from a book merely as pattern shows that
semantics can indeed be plausibly derived and extracted on the basis
of spacetime relationships in the environment of the cognitive
process. No magic or prior knowledge of linguistic meaning is
involved in bootstrapping the process.

\subsection{Concept identification from transverse stabilization of hubs (sleep phase)}\label{regionsec}

Post processing (what we might whimsically call `sleep maintenance'),
in between episodes of narrative learning, is used to mine the and
compare hubs. Fragments are now fixed, and what comes from their
repetition is merely a kind of frequency histogram. Hubs, on the other
hand, show patterns of co-activation on the episodic scale.  If one
assumes that similar patterns, supported by similar contexts, imply
similar concepts then one can begin to merge together contexts from
different experiences by looking for overlap between the hub fragments
(see figure \ref{coactivation}).  Without this, hubs would forever be
limited to their own episodes, and no lateral comparative thinking
could take place without an explicit learning episode. A `brain' that
can examine itself, measure and compare approximate degrees of overlap
between remembered contexts, therefore has the advantage of being able
to learn more from its learning by inference of similarity.

In going from hubs to connected regions of hubs, we move up one scale
in the hierarchy---a scale at which different narratives can interact.
One proposal for finding concepts would be to
seek out the stable regions from the principal eigenvector of the hub
graph---however that interpretation would leave concepts disconnected
from sensory input. A better interpretation is to attach concepts to
small fragments and consider these larger regions to be themes.

In order to measure our hypothetical concept formation, the overlap
sets between hubs belonging to proximity-connected regions was
assembled and summarized by their apparent content. This
is the analogous process to the longitudinal stabilization in section
\ref{long}. 

Based on the assumption that similar ideas might emerge from similar co-activations,
even where certain words are replaced---that could be the very mechanism by which
such impostors become effective synonyms. Thus, on the assumption that association
by coincidence is the start of everything to eventual meaning, with only pruning
by an observer's selection process remaining to form final distinctions.

\begin{figure}[ht]
\begin{center}
\includegraphics[width=7.5cm]{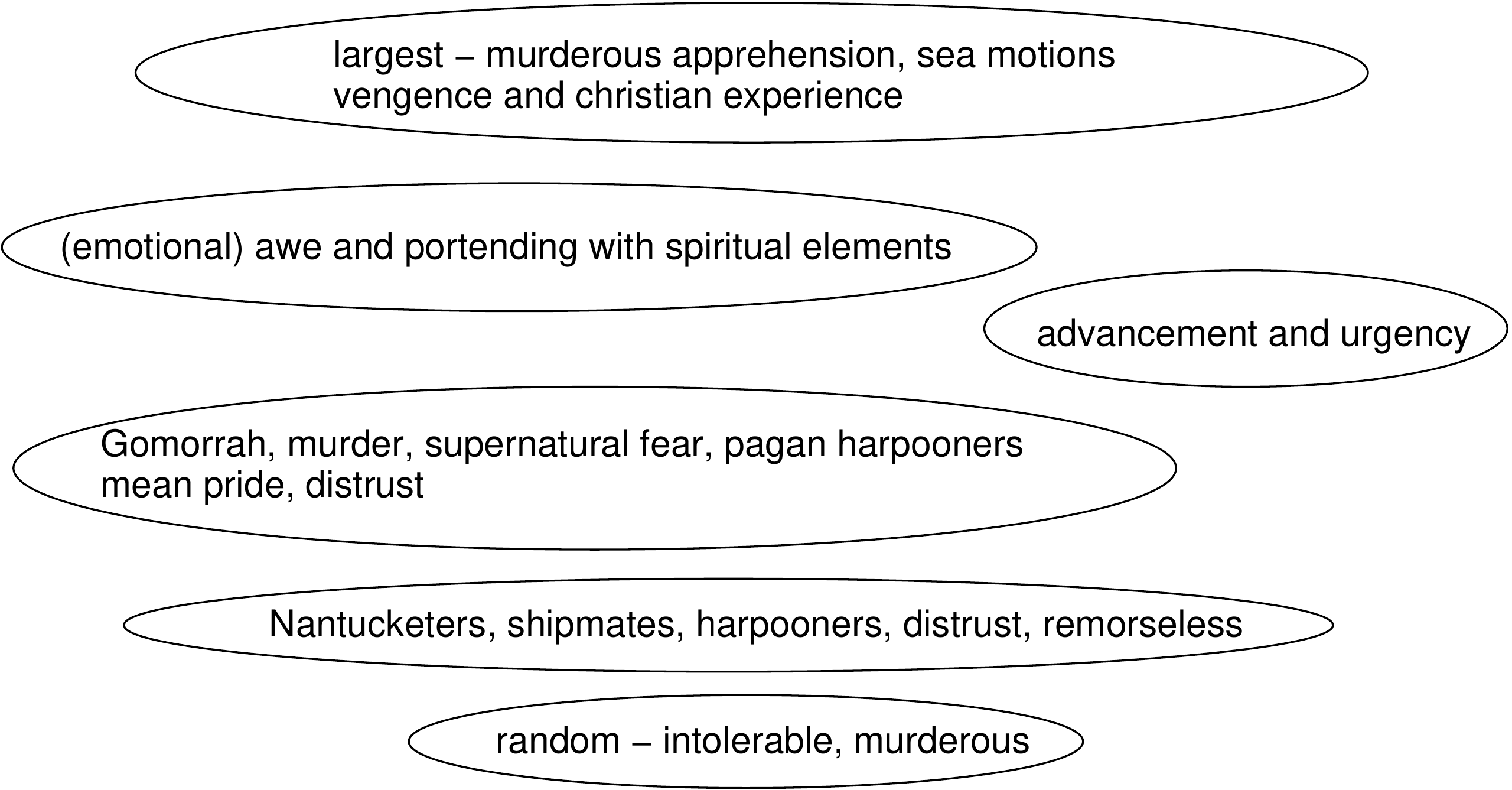}
\caption{\small The proto-emergent concepts from the novel Moby Dick (6 regions, from 298 links in 81 hubs).
On mixing with another sea novel, this structure simplified to just two regions, showing
how proto-concepts can be unstable to new learning.\label{hub_mobydick}}
\end{center}
\end{figure}

Making blind notes about what the mixtures of fragments might represent, leads to the 
annotation of regions shown in the example figures \ref{hub_darwin}-\ref{hub_mobydick}.
We can demonstrate some very simple cases, though the amount of information is far too great
to represent in a meaningful way, but we can indicate the apparent workings.

In the following sections, I've tried to sketch out how we understand
reasoning on each scale of a memory system. Unlike those who hold
logical reasoning to be the fundamental form, without a plausible
origin story, I wish to take a different
view\cite{cognitive,smartspacetime}, which is that reasoning is a
special case of storytelling. The fundamental process on which reason
is based is the stringing together of concepts and events until the
cognitive agent concerned reaches a satisfying emotional threshold.
During waking hours, this is connected with sensory input; during
offline processing there are no constraints\footnote{In dreams we feel
  that even bizarre behaviour is reasonable, even as our analytical
  brains question it, probably because the emotional sensation of
  resolution is triggered bringing a sense of satisfactory outcome.}.

\section{Geometry of reasoning in semantic spacetime}

In the final part of this study, I want to consider how to generate
narrative from what has been learnt by an agent. The narrative
hypothesis suggests that the ability to perceive one's surroundings and
imagine others that we haven't directly experienced must come from the
ability to recombine experiential data into new artificial
experiences.  To test this, let's first summarize the simple reasoning
based on the four spacetime semantics.

\subsection{Three scales of conceptual reason}

The principle of separation of scales leads to natural identification
of three qualitatively different scales. We can refer to these
as micro(scopic), meso(scopic), and macro(scopic).

Sensors sample episodic narrative on a mesoscopic scale
(sentences), which is then fragmented into microscopic fragments $\phi_n$,
with a lowest level alphabet $\phi_1$ (in this case words).
Aggregation of these fragments as `activation signals' into
macroscopic context hubs encodes the activation pathway from partially
overlapping context to sets of related memory events, so that---when
new episodes that contain similar fragments arise---the memory of past
related events will be activated by its semantic encoding (rather than
by a numerical lookup address, as in the lowest levels of a computer).

The containment hierarchy looks like this:

\begin{center}
\begin{tabular}{|l|l|l|}
\hline
Micro & Words     &  $\phi_1$\\
\hline
Meso  & Sentences &  $S_t \supset \{\phi_n\}$\\
\hline
Macro & Mixtures  &  $H_i \supset \{ S_t\} \supset \{\phi_n\}$\\
\hline
\end{tabular}
\end{center}

At each level, causal order information is preserved.
Fragments $\phi_n$ are essentially ordered sequences:

\beq
\phi_n \equiv \phi_1 \promise{\text{followed by}}\phi_1 \promise{\text{followed by}}\phi_1 \ldots
\eeq
Sentences $S_t$ are similarly capped fragments:
\beq
S_t \equiv \phi_1 \promise{\text{followed by}}\phi_1 \promise{\text{followed by}}\phi_1 \ldots
\eeq
Hubs express non-ordered aggregations of fragments; however, hubs are themselves
ordered by changing patterns of activation context, which bind an episode together.
So a narrative episode $N$ can be expressed on two scales, $N_S$ and $N_H$:
\beq
N_S &\equiv& S_t \promise{\text{followed by}} S_{t'} \promise{\text{followed by}} S_{t''} \ldots\\
N_S &\equiv& H_i \promise{\text{followed by}} H_j \promise{\text{followed by}} H_k \ldots\\
\text{where} & & H_i  \promise{\text{contains}} S_t, S_t'\ldots
\eeq

Context is a pool of recent patterns which gets accumulated by linking into hubs.
Relevance can be scored for fragments. Sentence relevance is scored as
the sum of relevances for its fragments, similarly for hubs.
Over time, fragments which are never reactivated would fade away, to be cleaned
up by garbage collection (another offline `sleep' function).

\subsection{Where are the concepts?}

The question of where concepts are within this system of information
seems subtle, and unexpected from a linguistic perspective.  One might
imagine that concepts have to be large aggregate structures with many
cross references: after all, our ability to have complex ideas seems
more sophisticated than simple sensory discriminators. However, this
appears to be incorrect. To be rooted in invariants, concepts have
their origin in fragments of the input language, whence more complex
and nuanced representations, on the scale of the concept language, can
develop from what we call the `themes' of the input language. Concepts may
ultimately become represented across several scales.

Consider the scales: context is a characterization of a cognitive
agent's current state of assessments of itself plus the input stream;
meanwhile, an event has to refer to changes in those states about the
agent and the exterior world, else it expresses nothing of concern to
the agent. The semantics of those changes thus map to the attributes
of concepts. The expression of concepts has to begin with the simplest
input invariants. Concepts must be smaller than events in order for
events to refer to them.  The genesis of the most basic concepts thus
appears to begin within the small fragments $\phi_n$---in the
spacetime phenomena of a cognitive agent's environment.  Similar
concepts might later be re-represented in other encoded forms, though
gratuitous recoding one-to-one would be wasteful and would serve no
purpose. Parsimony suggests that the concept language would refer to
different concepts than the input language, but the distance between
the two would remain short for their mutual constraints to be
effective\footnote{If the distance between the input language and the
  representation in concept language is short, a cognitive agent with
  a multiscale representation of sensory and recycled-sensory data
  would easily support several co-existing language representations.}.

What's interesting here is that proto-concepts must exist in the input
language itself. They may be embellished and aggregated by rescalings,
but we need to understand that process from the bottom up. 

\begin{enumerate}
\item A string of micro-concepts: e.g. phrases linked (fear, New York).
\item A string of meso-concepts: e.g. sentences linked (I once experience fear in New York. I walked to Brooklyn and found a dog.).
\item A string of macro-concepts: e.g. linked fragments (e.g. collectively suggesting the theme `fear loathing new york')
\end{enumerate}

As small fragments, input concepts thus behave like reusable encodings
or regular expressions\cite{lewis1}, which bind together in a process
of recombination ($\pm$ promises, in promise theoretic parlance).  A
sequential string of lock-key bindings could then trigger a cognitive
process, which ended with an outcome over some larger scale.  One can
therefore discuss whether a concept is the invariant trigger or the
dynamic process that unfolds from it. All non-local structures are
effectively processes, somewhat in the manner of a search algorithm.

For example, consider taking some random excepts from the input, and
using CAPS to represent a hypothetical association to `larger ideas'
on the concept language level:

\begin{center}
\small
\begin{tabular}{|c|l|l|}
\hline
 $n$ & IL $\phi_n$  & CL\\
\hline
 2 & sweating and panting & FEAR \\
 2 & utilized system          & WORKHORSE\\
 2 & whole coordination       & TOGETHER\\
 2 & without understanding    & PERPLEX\\
 3 & algorithmic behaviours generally & DISCIPLINE \\
\hline
\end{tabular}
\end{center}
How concepts combine, is only by sequence and gen-sequence. So the input fragments
`sweating panting spit curse coordination sky-towers' might end up mapping to
an effective phrase in the concept language `FEAR AND LOATHING IN NEW YORK' through
a sequence of vertical and lateral graphical transformations.
Not also that any correlation between the fragments and concept language is (of course)
imaginary and for the convenience of godlike observers only. We might learn
the significance over longer experience.

The promise theoretic basis of the Spacetime Hypothesis suggests
\cite{cognitive4} that input level concepts might be principally
identified with invariants of $\phi_2$ and $\phi_3$, owing the the
linear bindings of a timelike stream. Longer phrases could still be
technically significant due to the effective compounding of words, but
their reusability becomes decreasingly likely with longer phrases.
Indeed, the data show that there is close to zero repetition of
$\phi_n$ for $n>3$. This suggests that we look for concepts in the hub
fragments of $\phi_2$ and $\phi_3$, especially those which overlap
between different contexts. Those fragments will be the basis of a
concept language. Assigning names to those concepts becomes the
commentary language,

On further reflection by an agent, small fragmentary concepts could easily become
embellished with `bells and whistles' by superagent clustering of
fragments. This could occur by microscopic combination of the input language,
or by graphical structure in the concept language. The latter would occur by the insertion of new hubs, but these
could come from a sensory context without new information---they can
only come from the running context cache of the agent, i.e. `what it
is currently thinking about'. So hub formation by interior ruminations
is a context driven process that could be performed offline (e.g. sleep phase).

It seems plausible that concepts and themes may have similar
geometries but on different scales---and thus not truly independent
ideas, as they can always be transmuted into one another by scale
transformations\footnote{Renormalization probably plays a significant
  role in reasoning.}.  This is not to say that they are scale free
`fractal' representations. If one believed in a scale-free phenomenon,
there would be no reason why this process would stop, but the scale of
observations and sensory inputs is not without limit. Indeed, it's
pinned by the outside world of the observer, and that breaks the scale
invariance in the natural way that all symmetries are broken: by
boundary conditions.

Retaining multiple scales of pattern fragments (demarked by spaces) is
likely an unavoidable strategy to find the effective boundaries of
concept fragments.  Once themes have been rendered as invariants on a
larger aggregate scale, they are ready for recombination using the
same rules as for $\phi_n$ fragments, and the whole process can
potentially start all over again\footnote{How we humans manage the
  boundaries of a concept in a knowledge representation remains
  entirely unknown.} Would there be more super-hubs? Concepts have to
be decorated with contextual information, which is captured by hubs
(mixtures), but the same concept can also exist independently of a
very specific context. Boundaries seem fluid things, but the constraints
of limited resources must naturally prevent that from happening.

\subsection{What distinguishes proper names?}

\begin{figure}[ht]
\begin{center}
\includegraphics[width=6.5cm]{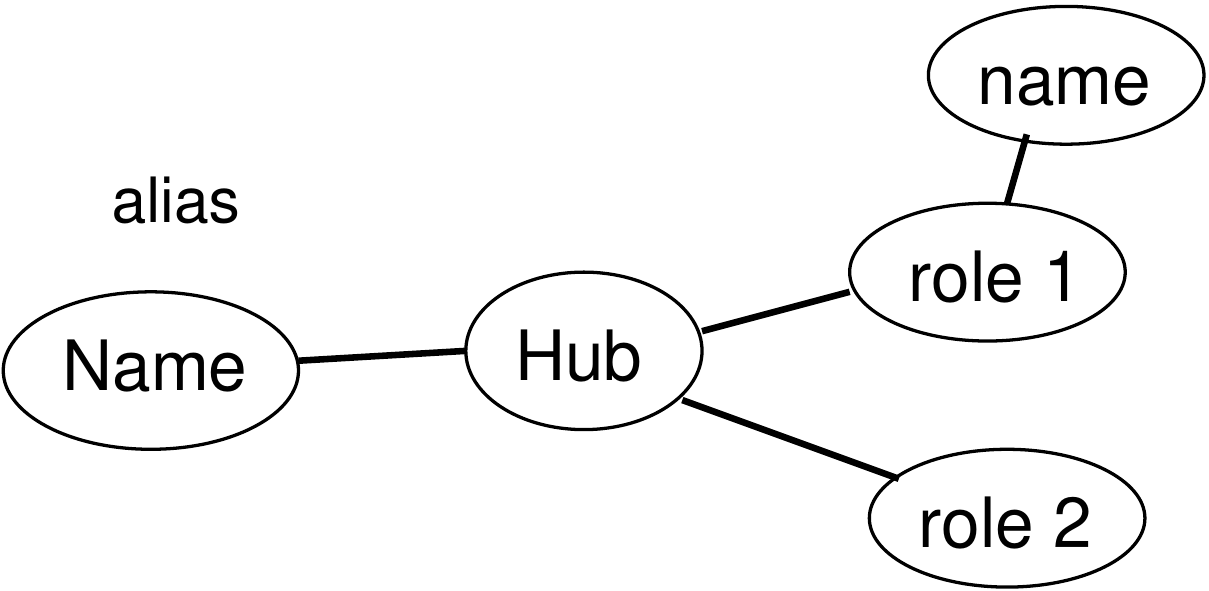}
\caption{\small Roles and names with their hubs. A proper name can be assigned to each collection of things
from the sum of the components. Aliases can be added later, in whatever language one needs, with a specialized
kind of link. This suggests that proper names might have a particular significance in memory networks,
in mapping between languages.\label{roles2}}
\end{center}
\end{figure}

A special kind of concept is a string that stands as its moniker: a
proper name.  Ultimately all representations in language hark back to
labels that are effectively proper names for something, and these
invariants become the favoured information expressed by concepts. Later, semantics become
altered.  Consider the idea of recording the name of person, e.g. John
Smith. In a traditional ontology, or relational database, one would
have separate labelled associations for the different attributes in
the record.
\begin{quote}
\small
Given name: John\\
Surname: Smith
\end{quote}
We can note that the surname in many cases is simply derived from a different source: the occupation
of the person in ancient times (Smith, Cobbler, Burgess, etc), or the village from whence the family came
(Jack of London, becoming Jack London, etc). So the semantics of surnames have evolved from being a role
to a qualifier. The same principle can be adopted to conjoin any kind of data. Indeed, the procedure is
formalized in relational databases by using join tables (see figure \ref{roles2}).

Names are thus part of a semantic coordinate system.  In order to
facilitate the addressability of data, by semantic lookup key (index
item) rather than by numerical coordinate in a Euclidean space.
Rather than keeping every combination of given and family name, one
can rationalize the findability by either full or partial name by
using the structure in which the full name unifies the component
names.  The the full name becomes a namespace for the partial names.
This principle can be applied to multi-dimensional names too, e.g.
street addresses, which have street, house number, district, region,
country, post code, etc.

At what scale might we expect to find structure that correspond to
`concepts' in the sense we understand in human thought?  According to
the rules of this study, the linguistic nature of the input data is
irrelevant, and we should not be swayed by our prior knowledge of
input language\footnote{Our inability to ignore or `unsee' grammar is
  a hindrance in identifying fragments, and believe in the results of
  this analysis.  There is an awkward compulsion to select things we
  understand and eliminate nonsensical fragments on the basis of
  understood usage, but this impulse has to be stifled and the data
  rigorously treated blindly until the moment of rightful
  comparison.}, because our artificial cognitive system has no
knowledge of language---it sees the data stream simply as patterns.
one pattern is as good as the next. So word fragments certainly can't
be significant enough to correspond to concepts, any more than codons
or single genes correspond to unique biological characteristics.
However, different admixtures of these will contribute to
characteristics. The question then is: will they be stable and
distinguishable, or spurious and prone to muddle?

The first question is: how should be identify clusters? Hubs are the features that can
aggregated the meaning from a number of stimuli. A full name hub like ``John Smith'' can
join together ``John'' and ``Smith''---but how do we understand these parts? How do we know
which is family name (role or context) and which is given name ``identity tag''?
Does the distinction matter? In order to distinguish name from role, we can only allow hubs
that have two nodes connected to them, so that the types are distinguishable. However,
the seems dynamically inefficient, and suggests a mechanism that would not evolve naturally (see figure \ref{roles}).

The key distinction between roles and names is that roles are repeated
to intentionally signify similarity, whereas names are only repeated
without the suggestion of being similar.  Another possibility (figure
\ref{roles2}) is that hubs may have proper names and any number of
roles, or simply names some of which are re-suable roles and some of
which are not identifiable as reusable roles.  Then, we have to take
into account the role of selection to prune the routes emerging from a
hub.  Some routes could be assigned weights stigmergically.

With a counterpoint in a selection process, the onus of identification
can be more on learning at multiple timescales.
\begin{itemize}
\item A name is a singleton node. A name could be a pattern pulled out of an event. Or it could be random.

\item A role is a contextually supported fan, because it is itself a hub formed from multiple sensory inputs.
\end{itemize}

A smart sensor would project data into a vector of semantic categories (a matroid or basis set). This vector
plays the role of the width freedom in a neural network. The weights become `polarized' by data from the
environment, so there is a kind of semantic `compass' implicit in this approach\cite{stories}.

\subsection{Hubs and their namespaces}

\begin{figure}[ht]
\begin{center}
\includegraphics[width=8cm]{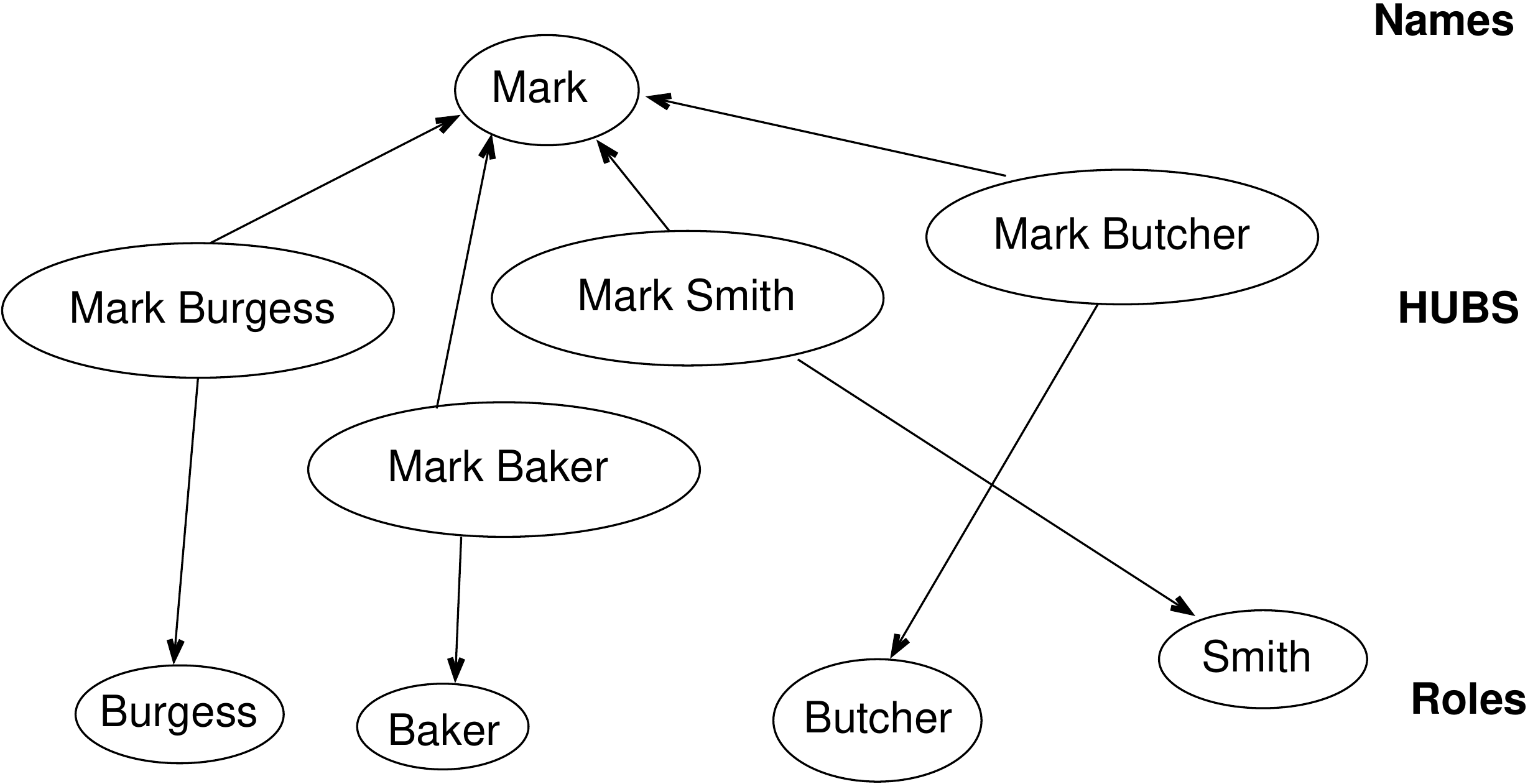}
\caption{\small Roles and names with their hubs. Simple hubs behave like full names
composed from given and family names, or name and role, name and address, etc. Thus when
hubs are not named with intent, they can emerge from their component fragments, and those
fragment mixtures spell the effective name of the namespace. Eventually, overlap
interferometry will separate and clarify namespaces into effective concepts at this level too.\label{roles}}
\end{center}
\end{figure}

Hubs are used to draw attention to the spanning sets of fragments. A
single hub connection uses the matroid promise
pattern\cite{promisebook} to conjoin all members of a set to single
unifying node, which can then be given a single new name to refer to
the entire set.  The principle can be understood on a small scale
before scaling it to arbitrary clusters (see figure \ref{roles}).

Namespaces constructed hierarchically in this manner may be quite fluid,
since the patches of members, which are referred to by the hubs, can
overlap and the boundaries between them can be rewritten in the light
of new experience\footnote{In relational database theory, the normal
  forms are usually taken as rules of thumb to avoid this kind of
  overlap, but the model is rigid and overconstrained. Hypertextual
  networks, for instance, may have many paths between items, with
  different interpretations.}.

Using a principle of aggregating nodes that belong together
(sentences, fragments, etc) under a single node which represents the
aggregate one can then form the name of the aggregate from the direct
sum of the parts. This is the approach used here. One advantage of the approach
is that it means the full name can be decomposed directly into its components very easily
to find when aggregate regions have common members.

Because hubs names are internally orderless, when different narratives
are merged, the hubs that are condensed from their input processes are
free to overlap, due to the presence of common fragments. This doesn't
happen naturally in a realtime sequence based on events, because the
process order creates overconstrained distinctions (as in logical
models).  Hubs that encode context during single episodes lead to
distinct silos of non-causally connected parts---each episode creates
new and unique context hubs. So hubs can never connect independent
narratives from the cognitive process itself. Member events
(sentences, in this case) will probably never coincide by accident,
since the probability of precise repetition of long sequences is
extremely low. 

This is why fractionation into small atomic constituents is the
mechanism by which parts can be decomposed into an elemental
chemistry.  Fractionation acts as a prism to split composite input
into distinct patterns, that may be divorced from context.  Thus, in
post-processing (sleep maintenance), compositional similarities
(overlapping spectra) can be recognized by sweeping through the graph
and linking hubs that are sufficiently close to one another (where
`sufficiently close' remains to be determined)\footnote{This search
  process is basically like `web crawling'.}. The fragments generate a
matroid basis for vector comparison.

Hubs with significantly similar spectra can be assessed as `close' or proximal
in their basis elements, regardless of their origin. The likeness to a
vector space can be made by approximation and large numbers. Thus we
may define a plausible notion of distance between hubs. This is the
significance of hubs---why we need a parallel representation of the
input language in a disordered state.

We therefore operate with two parallel representations of the input
patterns in this study: an ordered summary, which represents precise
contextualized recall, with causal order intact, and a disorder
spectral mixture which is conducive to inter-episode association.

\subsection{Generating narrative}

If we could generate new stories based on old ones, in a plausible
way, then we would have a simple model of reasoning based on past
learning.  If this could further be constructed from the four
spacetime relations then we would have plausible evidence in favour of
the Spacetime Hypothesis.  These artificial narratives could then be
assessed for their credibility: stringing fragments together is
easy---but they also have to make sense to a human arbiter.
This is a challenging demonstration because, while the data structure
is simple, the combinatoric search space is huge and the number of
overlapping possibilities that occur in parallel is daunting.

There is a set of transformations from the alphabet of words $\phi_1$
and their strings $\phi_n$ to sentences, which are complete constrained
statements: states, relations, or actions, to context accumulated
over recent past concerning all such statements, with a constant rate of
forgetting. Sentences are partially ordered events, and therefore must
retain causal order in episode encoding.
When making up or telling new stories, pathways labelled by the ``followed by'' relationship
offer possible `completions' that may score differently by relevance.
Generalizations are offered by the `containment' relationship.
The accumulation of pathways formed through these relations results in a simple geometry based
on the four semantic types. Scalar expression is encoded within names or
labels, and containment is built through hub matroids, causal order is a
simple chain relation $>$, and proximity is an undirected relation between
hubs.

Reasoning is sometimes associated with logic. Logic is just one form
of highly constrained narrative, which can be derived from a graph of
relationships. However, if we start from a graph of concepts, linked by the four spacetime relations,
paths through the graph correspond to stories to be told. Strictly speaking,
these are stories of the commentary language. However, in this work,
for the sake of conceptualization, we've identified concepts with
fragments of input language---trusting that our extraordinary ability
to think associatively and make sense of fragments, even when disordered and unruly,
will allow us to make recognizable stories from those fragments.
The result might not be grammatical, in the normal sense, but they will follow
a deeper kind of grammar, which---according to the Spacetime Hypothesis---is based
on the underlying spacetime process of sensory data gathering about the world.

Pathways through the graph are stories, and the graph grows somewhere
between linearly and quadratically in size.
The average number of possible stories per length of input stream data
therfore grows considerably. Some kind of selection process (call it an
algorithm) is needed to regularize spurious connections in a graph.
Many authors have tried to construct this using descriptive logics,
however logics are usually overconstrained and tend to result in either
nothing at all or just a one-to-one copy of the input.
The shortcomings of descriptive logic approaches were one of the contributing
elements to the development of Promise Theory and its long-standing
relationship with knowledge representation.

\subsection{Episodic order (causality)}

Our simplest and most common understanding of narrative is based on
linear storytelling: episodic recall of a stream, like playback.
Timelike events are joined together by precedence promises, creating
parallel fibres of narrative that are bounded by the start and end of
the input stream (here that means a document).  The criteria for
starting and stopping are important: as we've seen if episodes become
too long, the meaning of them may become clouded by a lack of
perceived focus.

This kind of linear lookup doesn't scale well for searching and dropping
into knowledge from different angles or requirements. This is why we
make tables of contents and indices in textbooks. The linear linguistic form is
s convenient as an interchange format for passing on ideas, but
it's not the way we think inside our heads. Look-up is based on a running
context of the observer's thought process, which is completely disconnected from
the original author's thought process. So we begin to flick through the book
looking for simple fragments to latch onto. From there, we might start reading
a little. Then we go to the section header or some summary of the particular
paragraph. All this is facilitated by the highly geometrical construction of books.
Later, those geometric aspects were generalized by hypertext, but they remain
essentially intact. A question that arises here is whether the chapter-by-chapter processing
of the Thinking in Promises book played a role in keeping its focused concepts 
under control, by effectively restarting a new learning experience in each chapter.
This certainly deserves further study, as `taking a break' could be a way in which a cognitive system
maintains order. We must defer that question for a later time.

The four spacetime semantic promise types allow structures that enable
simple narrative playback, indexing, titling, and cross referencing
without the need for an extensive ontology.

\subsection{Descriptive elaboration (scalar expression)}

Expression is a scalar promise, i.e an agent's promise about self.  It
refers to an interior property of the agent making it, e.g. state,
colour, height, name etc. Expressions are effectively the proper names
for conceptual properties; they are the way we elaborate on
descriptions (having the role of names, adverbs, and adjectives).
Although initially sceptical of this view, it agrees with the
orthodoxy as described by \cite{feldman1}, and doesn't exclude the
transmutation of larger themes into new concepts at the scale of the
concept language.  Properties expressed by fragments (on any scale)
seem to be indistinguishable from the role of a proper name. They are
scalar attributes, and thus play no role in relating agents to other
similar agents\footnote{Although one can represent a scalar as an
  artificial vector in extra-dimensions for the purpose of
  representing the matroid basis, that's a separate issue.}.

\subsection{Association by co-activation (containment)}

\begin{figure}[ht]
\begin{center}
\includegraphics[width=4cm]{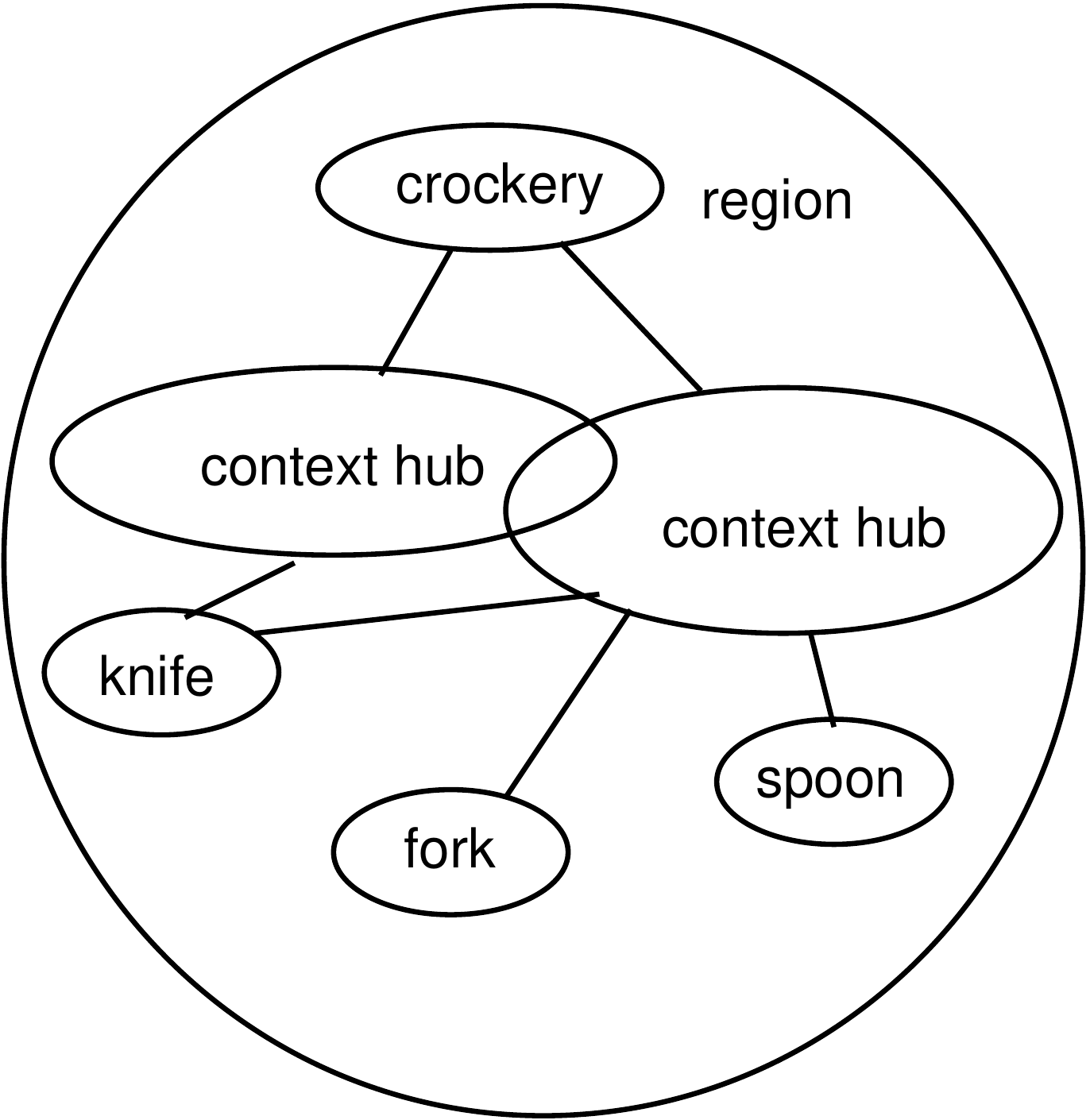}
\caption{\small The generalization of a concept in the input language
  to a higher level, with a name in the concept language occurs
  through fractionating context and identifying transverse invariants.
  What spells `crockery' in the concept language will in fact be an
  overlap spectrum of fragments between similar hubs that contribute
  to their support.\label{crockery}}
\end{center}
\end{figure}

If we consider our understanding of concepts, we assemble
smaller concepts into larger ones by generalization. 
This is a vertical aggregation in figure \ref{hubscales}.
Ideas like `walking', `running', `ambulation' might fall under the
larger idea of `movement'. How we arrived at those particular names
for the activities they represent is a long story. What matters is how
we emulate that origin story.
Similarly, one can imagine several contexts in which different eating implements
are represented. The overlap between these contexts leads to the combination
of a category (probably with some noise) in which knife, fork, and spoon are
represented strongly (see figure \ref{crockery}). The word `crockery'
cannot emerge from that process---whatever term is used there belongs to the
concept language, which has its own dynamics and etymology. We use the term
here for the convenience of readers with godlike powers of observation.

The analogue of this, in the present case, is how experiences with
similar contexts are collected under a common umbrella, given that we
don't have names for the terms $\phi_n$ in our own commentary language
(we have only the text fragments in the input language). We also can't
assign a meaningful name to the category they form. A simple way
around this is to take our cue from DNA. Fragments can be strung
together as a single chain, with punctuation markers in between
distinct members. When we need something from it, the chain can be
broken up into fragments again in order to be compared with receptors
for the specific patterns. This is like club membership by club and by member.

For example, a reasonable `proper name' for one concept derived
from the member fragments would be:
\begin{quote}
\small
1:responsibility, 3:this approach impractical,4:transmission does
not scale,4:scope of every transmission,5:were considered
immutable and unique,1:recipients,1:mechanism
2:before terminating,2:protected broadcast,3:internet
protocol provides,3:imposition protocol only,4:the
internet protocol provides,5:push-based imposition
protocol only half,2:protocol provides,3:for emergent
delivery,3:computer operating systems,4:scale
independent the assumption,1:configuration,2:prefix
\end{quote}
As a single large string, the name is not very memorable (for a human anyway), but it's functional---as with DNA.
It could be exchanged for an alias in the commentary language later.
We would likely agree that this is not a fully formed and well-rounded
concept, of the sophistication we are used to as human observers.
However, it is a valid proto-concept, ready to be joined to others
forming a larger region with greater focus and depth of contextualization.

A graph is multi-dimensional at every point.  We reduce dimensionality
to linear trajectories to avoid missing anything. In one dimension,
you have to bump into everything---but the embedding is still useful
for navigation.  The two-dimensionality (or rather two-typedness) of
spacetime relationships (membership in space, versus order in time) that
plays an important role in reasoning because membership in larger categories
plays a role in stories and reasoning is a form of storytelling.  We can take
a silly example:
\begin{quote}
\small
`Frag1' impacts `Frag2' and therefore affects club `HubA' 
of which `Frag2' is a member. Because the club pays its workers,
another one of its workers `Frag24' was laid off and this caused 
`Frag 26' to scream.
\end{quote}
The specific words in these relationships express attributes of the geometry
to add colour, but the essence lies in the geometry:
\beq
\text{causes} \rightarrow, \text{generalizes} \uparrow, \text{similar to},\text{e.g.} \downarrow, \text{causes} \rightarrow
\eeq
We sometimes infer causal relationships through intermediary concepts by generalization.
If an agent is a part of a superagent collective, then there is a sense in which the
superagent is responsible for the agent's own promises.
This is how scaling works in an agent model, such as Promise Theory.

\subsection{Association by similarity (hub proximity)}

When hubs are similar, the kind of vertical-horizontal reasoning
exemplified in the previous section can also pass from hub to hub.
One of the consequences of this cross-labelled geometry and the
implicitness of meaning is that mistakes based on spurious overlaps
can lead to new causal connections too.  For example, consider the
following scenario.  

\begin{quote}
\small 
The name ``Godzilla'' is the name of a movie and the name of a Maki
platter at a Japanese restaurant.  In an apparent Denial Of Service
attack on its booking service, a Japanese restaurant's website is hit
by a large number of requests, which brings it to its knees.  The
cause was a misunderstanding relating to a film promotion for the
upcoming movie Godzilla. The only connection between the two is the
name of one movie being shown, which has a promotional website of
similar name. A simple typing mistake is what leads to a very
different causal sequence of events. 
\end{quote}
Reasoning about this story
requires one to make a leap of contexts.  A concept was transmuted
into another by a `resonance' perhaps around a single proper name.
Associations can easily be made by proper name, functioning as a simple
semantic address.

As in all relativistic scenarios, one process compares itself to
another process, according to the rules of (+) and (-) promises, seeking a limited overlap.
Typically, the (-) process acts as a coordinate system against which
the (+) source process offers data, providing a calibrated scale for
measurement.  The covariant meaning of the overlap is thus judged by
each receiver independently. This is the meaning of relativity.

How might a word like `concept' become close to a word like `meaning'
or `semantics'? This could not happen in the input language---only in
the concept representation, because the input language can't be
measured semantically. The only possibility is that---over
time---co-activations encode associations on a higher level by
context. That could be encoded as hubs and regions into a concept
language representation of the ideas, which then overlap. 
Thus, at some moment in the history of the agent, these terms would
have to appear within the same co-activation cluster, or words related
to them would have to appear in the same co-activation cluster. 

A synonym in the concept language is thus a process based on an input
fragment that plays the same role in a reasoning process.  It could
apply on the level of concepts or on the level of fragments. However,
without some horizon or limit on the minimum degree of overlap, it's
potentially possible for many if not all concepts to be considered
close together\footnote{Counterfactual evidence may also add a further
  selection criterion for filtering clusters that have become too
  enmeshed in each others associations\cite{pearl3}.  That subject
  goes beyond the scope of the current work.}.

\begin{figure*}[ht]
\begin{center}
\includegraphics[width=16.5cm]{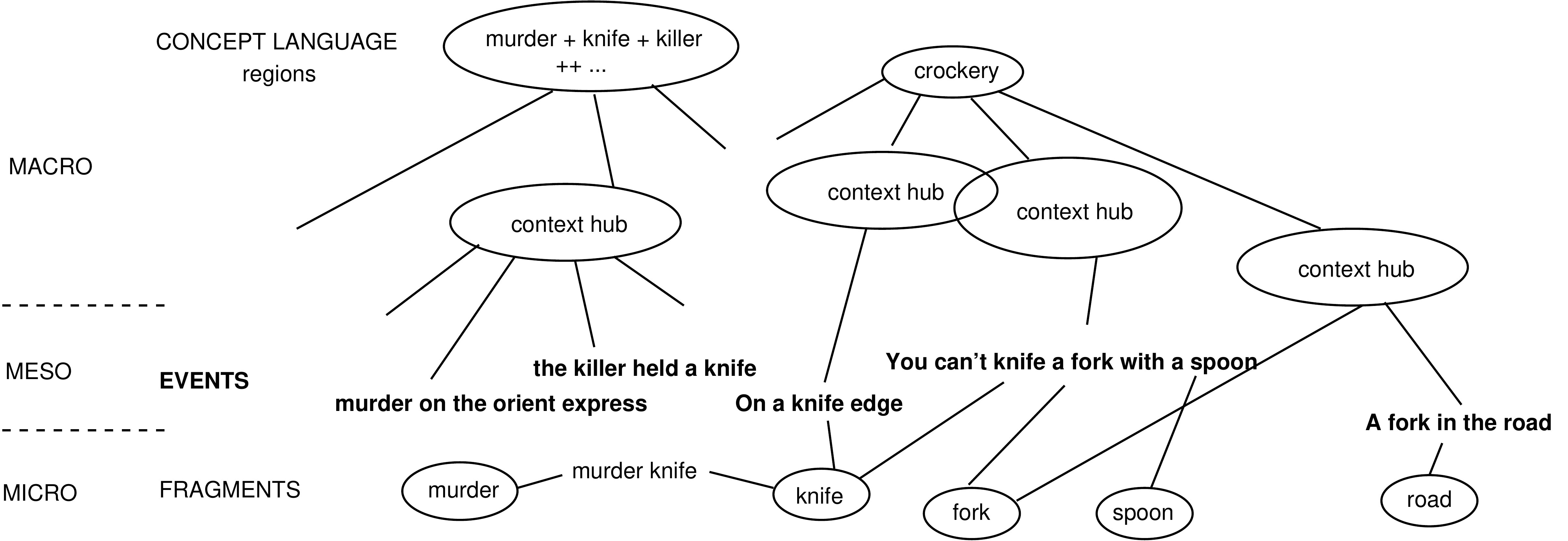}
\caption{\small Recombination can occur at the microscopic level, e.g. `murder + knife' which is purely
syntactic, or it can occur on the macroscopic region level by sufficient overlap of hubs. The macroscopic
concept is must more vaguely related to actual input phenomena, but corresponds more closely to
nuanced human ideas. When recombining these fragments on different scales, we can't expect the combinations
to lead to grammatically proper sentences of the input language, but we may use the labelling in the
input language to relate the inferences to phenomena at the sensory level.\label{murderbyknife}}
\end{center}
\end{figure*}

\subsection{Microscopic reasoning}

The most primitive level of reasoning is that which occurs in simple
single-scale processes, such as biochemistry---fragment recombination.
Recombination of patterns, on the microscopic level, involves
rearranging the words of the input language to elaborate its chemistry
and seek out new combinations, leading to new process outcomes.
Patterns will eventually be selected by their niche semantics: if they
bind to something and advance a process then they can become new process
invariants. For example: words `knife', `murder', can be combined as new
phrases:
\begin{quote}
\small
\noindent murder knife\\
murder by knife\\
knife by murder\\
murder contains knife\\
knife contains murder\\
knife leads to murder
\end{quote}
These are microscopic recombinations at the level of `utterances',
on the input language scale.  There are no rules for syntax
or grammar here, only rules for binding, precedence (follows), or
similarity (proximity).

From here one can pursue two approaches for generating new narrative
events: one based on proximity in the space of fragments, or one based
on causal order (time).  From our knowledge of statistics in paper 1,
and from the earlier discussion in section \ref{geom}, we expect the
different phrases will play different roles (see table \ref{table2}).
At the microscopic level, simple promises expressed by the
fragments, in whatever memory representation they find a home, will
constrain the way fragments can combine. This is surely the
origin of grammar.  Fragments can simply by combined in all ways
to create an associative binding (see figure \ref{murderbyknife}).

\begin{itemize}
\item 1-phrases ($\phi_1$) are like codons (component chemicals). They can overlap as atomic parts of any string.

\begin{quote}
\small
~\\ 
animals\\
hate\\
bananas\\
murder\\
knife\\
\end{quote}

These phrases can be combined into new phrases of higher order in $n$, e.g. hate bananas,
bananas murder, murder knife bananas, bananas hate knife murder, etc (see figure \ref{murderbyknife}).
We can't and indeed shouldn't expect such recombinations to resemble English grammatical
sentences in a normal sense of containing glue words (e.g. `murder by knife'), which are
probably habitual adaptations that normalize over long times,
but we may expect them to match those concepts in a more clumsily expressed form.

\item 2- and 3-phrases ($\phi_2,\phi_3$) are order-constrained combinatoric parts (may form mixtures).

\begin{quote}
\small
~\\ 
animals hate bananas\\
animals are big\\
animals are small\\
murder by knife
\end{quote}

These fragments already contain normalized binding words. If they repeat often enough they
could override the clumsier recombinations of 1-phases. The longer a phrase, the more
significant it is, and the less likely it is to combine further.

\item Longer $\phi_n$, together with sentences, are strongly ordered contextualized examples. These effectively
become playback events as the likelihood of them recombining is infinitesimal.

\begin{quote}
\small
~\\ 
infer that our domestic animals\\
domestic animals and the fact\\
animals now in\\
choice animals would thus\\
four-footed animals\\
domestic animals were originally chosen\\
that if all the animals\\
animals are now annually shot\\
the plants and animals which\\
liable and such choice animals\\
domestic plants or animals\\
\end{quote}

\end{itemize}
To summarize, small fragments $\phi_n$ ($n < 4$) can be spliced
together to recombine into artificial sentence events, corresponding
to new sensory experiences.  This is the simplest way in which new
expressions can be formed from old, or very elementary stories be
told, like film editing. This might not generate perfect phrases in so
complex a language as English, but they would not be difficult to
understand\footnote{In the complexities of English language, the glue
  words that join fragments together doubtless play a role in the
  emergence of a grammar that we would recognize. In other human
  languages, such issues are not relevant.}.  Following the causal
links between ordered word fragments, in this way, is one way to tell
story fragments. It's a form of playback of recorded experience, as
there are no causal links between different episodes, without
observing patterns of co-activation on a larger scale. Context hubs
allow microscopic stories to be routed across narratives. Thus
story-telling at the microscopic level is either unimaginative or
discontinuous.

\subsection{Mesoscopic reasoning}

Moving to the mesoscopic level, we may consider complete events as the basic
constituents of narrative: sentences and their ordered relationships to one another.
Finding relationships between events outside of their original context
is much more challenging, and thus far less likely to find a match that makes
obvious sense on the level of a single sensory episode. Thus, the nature
of mesoscopic narrative could be quite different to that on the microscopic
level.

How or where one begins a story based on events is an issue in its own right
that we probably can't answer fully here. The most natural way to bootstrap
a starting point would be to base it on running context.  Certain events
may stand out because the namespaces they belong to are `addressed' by the
buffer cache of running context. There can still be many possible starting
points, which leads to two possibilities:
\begin{itemize}
\item A random (non-deterministic) selection from the possibilities.
\item Several starting points are retained in parallel (superposition) and considered
alongside one another.
\end{itemize}

As an example, consider thoughts about the short fragment concept of ``animals''. As
a single fragment this can easily overlap with many contexts and events. So there
may be many possible events in play. There are two binding relationships in play:
for space and time:

\begin{itemize}
\item Space (proximity): Searching for parallel events containing this
  concept, one starts with hubs. We look for hubs that contain the the
  concept fragments, which takes us up the network layers from edge to
  a `central routing plane' (see figure \ref{hubscales}). Context looks something like this:
\bigskip
\begin{quote}
\small
2:the consciousness,2:the darkness,2:the individual
2:the kaleidoscope,2:the metropolis,2:the stationary
2:thinking willing,2:this birds-eye,3:across half-charted oceans
3:acting living carried,3:aspect--he was by,3:birds-eye aspect--he
was,3:caught his breath,3:distance close beneath,3:drinking
at water-holes,3:emotions were legible,3:everywhere
all-at-once dont,3:experienced the sense,3:experiences
of strange,3:extends the consciousness,3:family
presented themselves,3:fighting toiling loving
3:his dizzy elevation,3:hovering in mid-air
3:munching sugarcane while,4:courage
of the fly--he,4:craters flying above cities,4:creating
and destroying differing,4:darted across half-charted
oceans,4:denied partially at least,4:element of
air without,4:everything that compressed life,4:experienced
the sense such,4:experiences of strange distant
4:imagination figured this glorious
4:more intelligent than animals,4:movement and
singing when,4:music caught his heart,4:never could
articulately clothe,4:new method of communication
4:realised that birds had,4:realised--must
some day produce,4:rhythm movement and singing,4:secret
and mysterious life,4:separate objects definite
divisions,5:burst into colour rhythm movement,5:by the southern
sun intoxicated,5:carelessly carrying nothing
with them,5:colour heat light and beauty,5:colour
rhythm movement and singing
\end{quote}
\bigskip This is only a small part of an actual context, even working
at the level of compression in paper 1 on a single source stream,
making it hard to convey what's going on in these experiments.
Nevertheless, we try.  Having found starting points in hubs, one finds
candidate events linked to them, which score for relevance in terms of
their overlap with the running context of the search agent. For instance:

\bigskip
\begin{quote}
\small
``Thus, to return to our imaginary illustration of the
flying-fish, it does not seem probable that fishes
capable of true flight would have been developed
under many subordinate forms, for taking prey of many
kinds in many ways, on the land and in the water, until
their organs of flight had come to a high stage of perfection
 so as to have given them a decided advantage over other
{\bf animals} in the battle for life.''\\
ALTERNATIVELY (SIMILAR)\\
``With hermaphrodite organisms which cross only occasionally
 and likewise with {\bf animals} which unite for each birth
 but which wander little and can increase at a rapid
rate, a new and improved variety might be quickly formed
on any one spot, and might there maintain itself in
a body and afterward spread, so that the individuals
of the new variety would chiefly cross together.''\\
ALTERNATIVELY (SIMILAR)\\
``Seeing, for instance, that the oldest known mammals
 reptiles, and fishes strictly belong to their proper
classes, though some of these old forms are in a slight
degree less distinct from each other than are the
typical members of the same groups at the present
day, it would be vain to look for {\bf animals} having the
common embryological character of the Vertebrata
 until beds rich in fossils are discovered far beneath
the lowest Cambrian strata--a discovery of which
the chance is small.''
\end{quote}
\bigskip

Having found a hub that contains examples relevant to the concept,
from here one can go in any direction to find events or similar hubs
where the fragment reappears. This is somewhat like a simple text
search.  The resulting parts can be strung together into a story.  Not
all stories read like Hans Christian Andersen---lacking data
continuity, some are quite disjointed (e.g. which might explain the oddly
disconnected nature of dreams, which are unanchored by sensory context).

\item Time (precedence and playback) Because of the siloing issue,
  narratives tend not to propagate across episodes. Once inside an
  event stream a search will tend to lead to playback of episodic
  memories rather than innovative recombination. In other words,
  timelike recall is distinct from the lateral spacelike recall above:
  more deterministic. For instance:

\bigskip
\begin{quote}
\small
``Thus, to return to our imaginary illustration of the
    flying-fish, it does not seem probable that fishes
    capable of true flight would have been developed
    under many subordinate forms, for taking prey of many
    kinds in many ways, on the land and in the water, until
    their organs of flight had come to a high stage of perfection
     so as to have given them a decided advantage over other
    {\bf animals} in the battle for life.''\\
FOLLOWED BY\\
``this is scarcely ever possible, and we are forced
to look to other species and genera of the same group
 that is to the collateral descendants from the same
parent-form, in order to see what gradations are possible
 and for the chance of some gradations having been
transmitted in an unaltered or little altered condition.''\\
FOLLOWED BY\\
``when we bear in mind how small the number of all
    living forms must be in comparison with those which
    have become extinct, the difficulty ceases to be very
    great in believing that natural selection may have
    converted the simple apparatus of an optic nerve
     coated with pigment and invested by transparent
    membrane, into an optical instrument as perfect as
    is possessed by any member of the Articulata class.''\\
...
\end{quote}
\bigskip

\end{itemize}
Each new event followed modifies running context and alters
the search affinity of new memories.

The search for a concept fragment as short as `animal' naturally leads
to a wealth of matches that are too numerous to document here.  The
search space can be reduced factorially by taking longer fragments.
For example, seeing the many contexts in which the search term arises,
one might choose to refine the search by selecting qualifiers and a
larger fragment.  If we combine the fragments to make {\em less
  general} concepts of the input language, we indeed find more
specificity and fewer cases.  The same principle would likely apply at
the level of the concept language too, but we can't study that here.
Longer fragments that are activated by the search term include, for instance:

\bigskip
\begin{quote}
  \small ``Thus, to return to our imaginary illustration of the
  flying-fish, it does not seem probable that fishes capable of true
  flight would have been developed under many subordinate forms, for
  taking prey of many kinds in many ways, on the land and in the
  water, until their organs of flight had come to a high stage of
  perfection so as to have given them a decided advantage over other
  animals in the battle for life.''\\
~\\
``As we may infer that our domestic animals were originally chosen by
uncivilized man because they were useful and because they bred readily
under confinement...''\\
~\\
``for if all the marine animals now living in Europe and all those that
lived in Europe during the Pleistocene period a very remote period as
measured by years...''\\
~\\
``It is a truly wonderful fact--the wonder of which we are apt to
overlook from familiarity--that all animals and all plants throughout
all time and space should be related to each other in groups...forming sub-families
families, orders, sub-classes, and classes.''
\end{quote}
\bigskip

How should we choose between these? The obvious way to rank them is
by degree of overlap with current running context, modified along the way. However, on the
scale of this experiment and probably beyond, that doesn't necessarily lead to a clear
selection criterion. The degree of overlap between alternatives is bound
to be comparable for a wide range of alternatives that remain in
play or `superposed' in the process.
Suppose then we choose to look at a longer concept of the input
stream: `four-footed animals'. This pursues pathways that lead to
several possibilities, all of which are superposed or coactive, `in
play'. For example:

\begin{quote}
four-footed animals on the ground\\
four-footed animals on\\
desire and four-footed animals\\
\end{quote}

For longer strings, matching events precisely is easy.
This natural behaviour suggests a principle at work, which supports
the notion of specificity from paper 1.  A new hypothesis might
propose that the length, in terms of the input language, of a fragment
corresponds to a concept of a certain scale. Its specificity is
inversely proportional to its length.
The connection between conceptual specificity and importance in paper
1 is certainly evident for all to see: even relatively short longer fragments are effectively
unique owing to the size of the extended word alphabet.  At the sentence scale, we now
find a specific playback events by direct fragment
activation\footnote{This study is is unusual in that it retains
  complete sentences as events, and reproduces these as short episodic
  occurrences, in full.  This generates an illusion of proper grammar
  and sophistication. This is a deliberate illusion: we fall foul of
  such simple legerdemain in our daily dealings with memory and
  experience. It's probably how an artificial process will eventually
  pass a Turing test. This method is effectively used in all deep fake
  technology.}:

\begin{quote}
\small
The details of the room could be inserted later according
to judgment and desire, and {\bf four-footed animals} on
the ground might also discover later the point of
view of birds who, from a high altitude in the air, saw
everything at once."
\end{quote}
This doesn't form an obvious story connection on the level of English,
but remember that this is input language, and the language of raw
sensory streams may not be as familiar as we expect. Senses
exhibit a tendency to `see what we want to see' rather than what's
there, indicating that narrative could play with perception directly
owing to the spurious connections in semantically addressed memory.

Promise Theory predicts that selection plays as active a role as
source diversity.  An agent engaged in active cognitive processes has
its running context at all times. This is the context that would be
`saved' as a hub for activation. The degree of overlap between this
running context and past context is thus a measure of relevance, in
the cardinality of the fragments. Searching by proximity to the term
of interest, one can therefore identify fragments that are measurably
relevant: This is a form of lateral thinking based on simple
proximity.
Replay of these events is strictly causal (playback) story about four footed animals is
neither a smooth segue, nor does it lead further: the event is a
dead end within the present episode silo, so now---unless we abandon the
selection according to direct relevance---we must give up.

We may, however, still proceed through other context hubs, by asking
the question: what other concepts were in play alongside the
`four-footed animals' within the running context, and do {\em they}
match the present running context? Are there any similar hubs where the fragments
co-activate with other concepts that recently were active in running
context? At this point, associative reasoning may lose focus on
animals altogether.  Likely, searches need
superpositions of many of these fragments to make strong connections
under constraint. 

The lesson of this exercise is that it's very difficult to work one's
way out the silo of a narrative episode by causal reasoning alone.  The walls of
separation are high. This is why fragmentation is key.  Without the
anchor of sensory validation, stories told at a higher level perhaps
need some self-consistency in a string of concepts and experiences to
reach an `emotional resolution'.  It's not clear at this stage how
that might happen, since the representation of emotional triggers is
too primitive in the current analysis.

\subsection{Macroscopic reasoning}

At the topmost scale of reason, we have themes, represented by regions
of proximate hubs.  This overlap between context encodings, or
mixtures of conceptual fragments, are where we expect to find broader
themes as mixtures of several concepts (figure \ref{overlap_context}).
Since regions can be assigned proper names, and treated as short
fragments on a larger scale (e.g. short fragments of the concept
language that refer to longer mixtures of input language concepts),
themes can easily be renormalized as concepts on a new level and the
whole method of stringing together stories based on these can resume
on a larger scale.  However, now the number of concepts is far fewer
and more sparse than before, so this might take some leaps of faith to
comprehend their meanings, and the results will be correspondingly more
vague.

\begin{figure}[ht]
\begin{center}
\includegraphics[width=8cm]{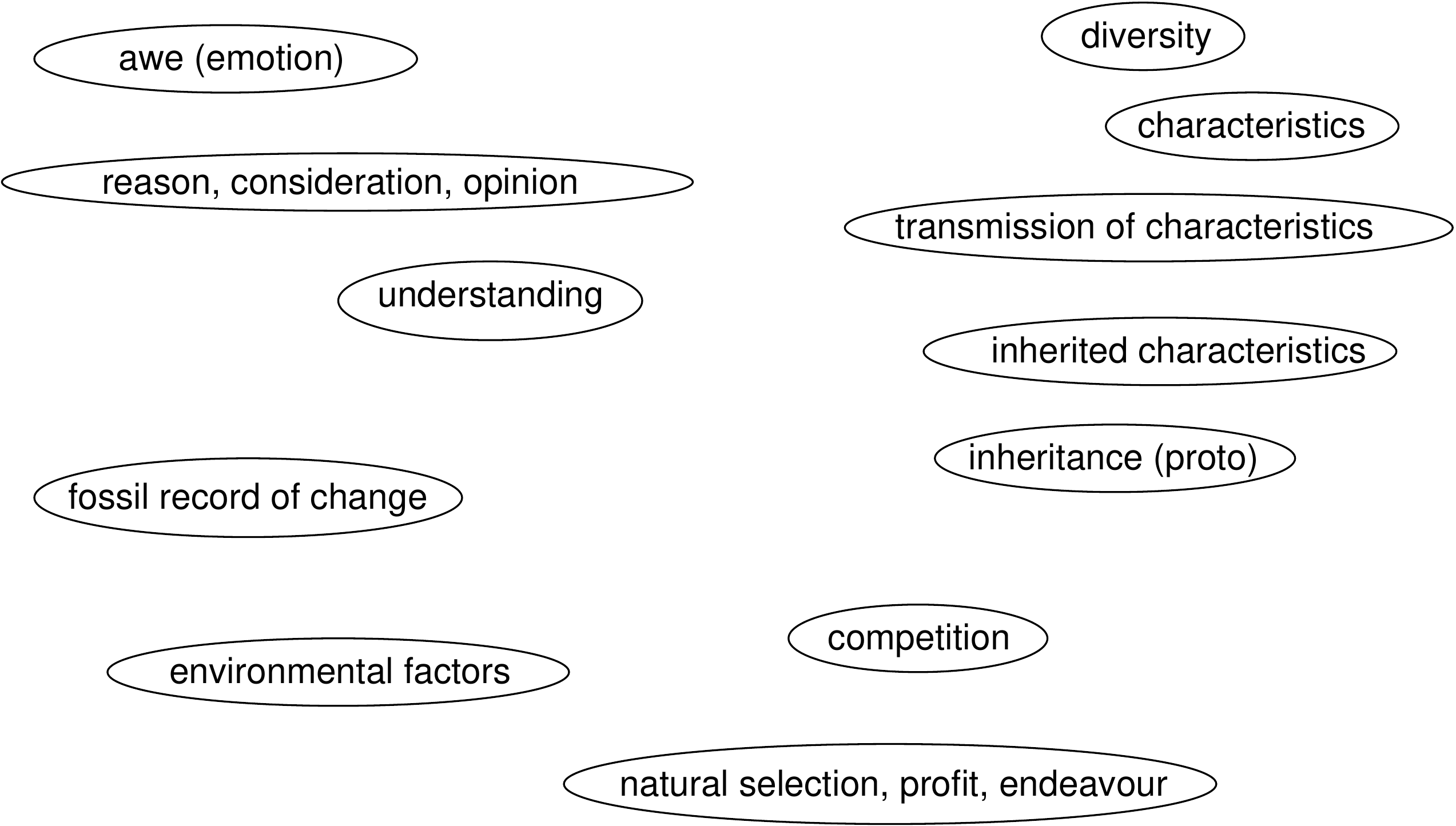}
\caption{\small The proto-emergent concepts from Darwin's origin of
species (12 regions, from 18 links in 70 hubs). The major regions seem
to revolve around themes of awe (emotional), considered reason, and
understanding. Then other regions concern the diversity of
characteristics in the environment, the fossil record in particular,
and question of how characteristics are passed on.\label{hub_darwin}}
\end{center}
\end{figure}

In Darwin's {\em Origin of Species}, the major hub-regions seem to revolve
around themes of awe (emotional), considered reason, and understanding
(see figure \ref{hub_darwin}).  Then other regions concern the
diversity of characteristics in the environment, the fossil record in
particular, and question of how characteristics are passed on. This is
not a bad summary of a well-known book. These macroscopic themes can be
interpreted as effective concepts at the concept language level.

Notice how this coalesced summary of the regions is not expressible by
the list of fragments that give the regions support from below. The
latter contains many more relevant details, but only in the input
language. The regions shown correspond to what could become concepts
in an eventual commentary language.

\begin{figure}[ht]
\begin{center}
\includegraphics[width=7cm]{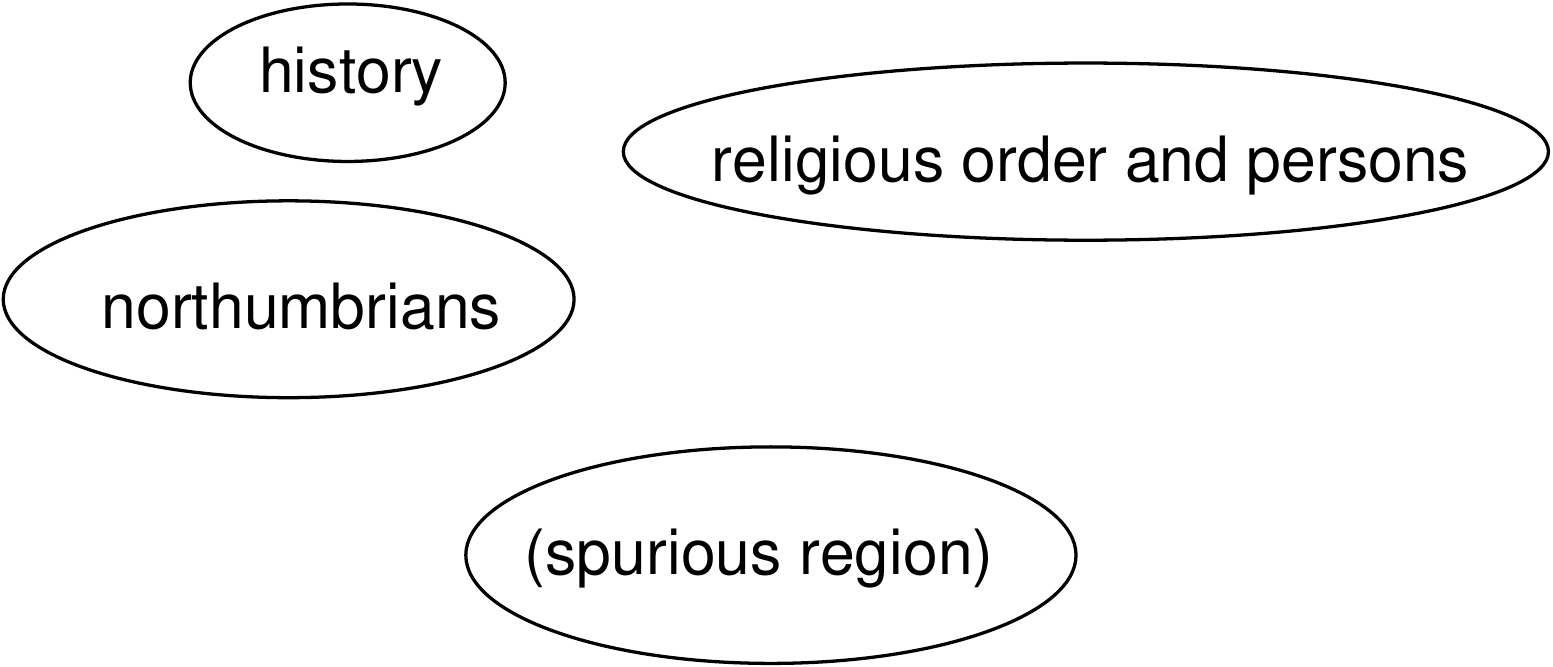}
\caption{\small The proto-emergent concepts from Bede's history (4 regions, from 7 links in 43 hubs). The major stable regions
are few in number, perhaps because the is little repetition of a small number of themes. The concepts revolve around
history, religious organizations, and Northumbrians.\label{hub_bede}}
\end{center}
\end{figure}

Given so few themes, or macro-symbols, the only narratives one could
generate from so few components would take the form of short strings
equivalent to English phrases.  This is reasonable: the more general
the concepts, the shorter our statements about them tend to be.  For
instance:

\begin{quote}
\small
There is awesome diversity of inherited characteristics in the fossil record or environment.\\
There is competition and natural selection in the fossil record or environment.
\end{quote}
For the history of Bede, the major stable regions are few in number, perhaps because the is
little repetition of a small number of themes. The concepts revolve
around history, religious organizations, and Northumbrians---i.e.
people from Northumbria, a region in the North of England (see figure
\ref{hub_bede}). The macro narratives take the form:
\begin{quote}
\small
History, religious order or persons as Northumbrians.
\end{quote}

\begin{figure}[ht]
\begin{center}
\includegraphics[width=8cm]{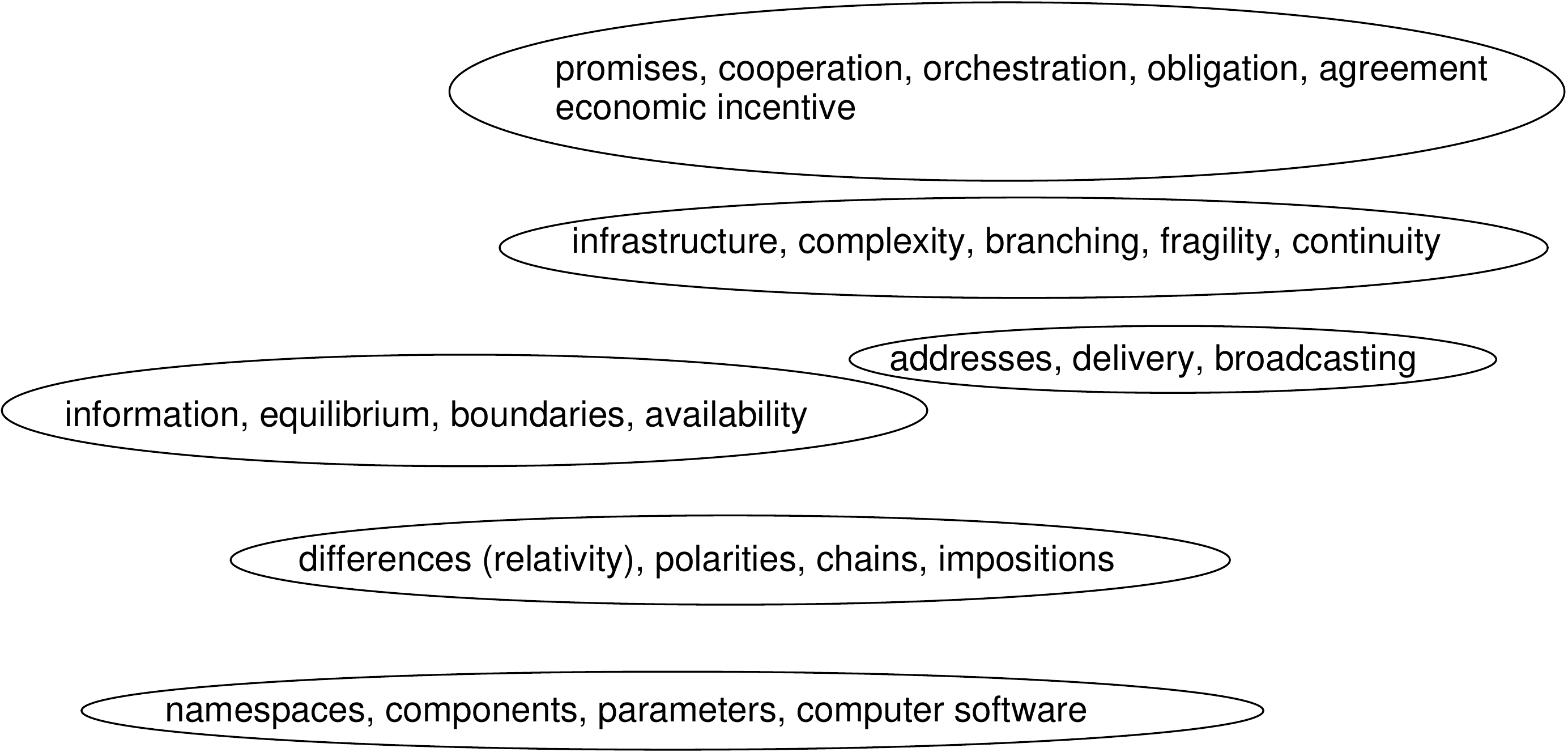}
\caption{\small The proto-emergent concepts from Thinking In Promises (6 regions from 14 links in 25 hubs).
The separation of concepts is not as clear at the level of regions, but the regions are decent
representations of conceptual themes in the book: promises and cooperation, infrastructure and its
properties, information and boundaries, impositions, and namespaces.\label{hub_pt}}
\end{center}
\end{figure}

For the {\em Thinking in Promises} book, the separation of themes is
not as fine-grained as a discerning reader would like, at the level of
regions, but the fragments within are nevertheless decent representations of
conceptual themes in the book: promises and cooperation,
infrastructure and its properties, information and boundaries,
impositions, and namespaces. What is interesting is the way the
concepts cluster together. The clustering does lump together related
ideas, and separate different ideas, given my own understanding as the
author. So, what we can say, from this little evidence is that the
outcome is not inconsistent with the intentions of the book. It's hard
to see how one could ask for more at this experimental proof of
concept level (see figure \ref{hub_pt}).  The macro narrative might
contain:
\begin{quote}
\small
Promises, cooperation, etc in infrastructure\\
Equilibrium and boundary in infrastructure\\
Equilibrium and boundary in promises, cooperation\\
Namespaces and information\\
etc.
\end{quote}

\subsection{Mixed narratives}

\begin{figure}[ht]
\begin{center}
\includegraphics[width=7.5cm]{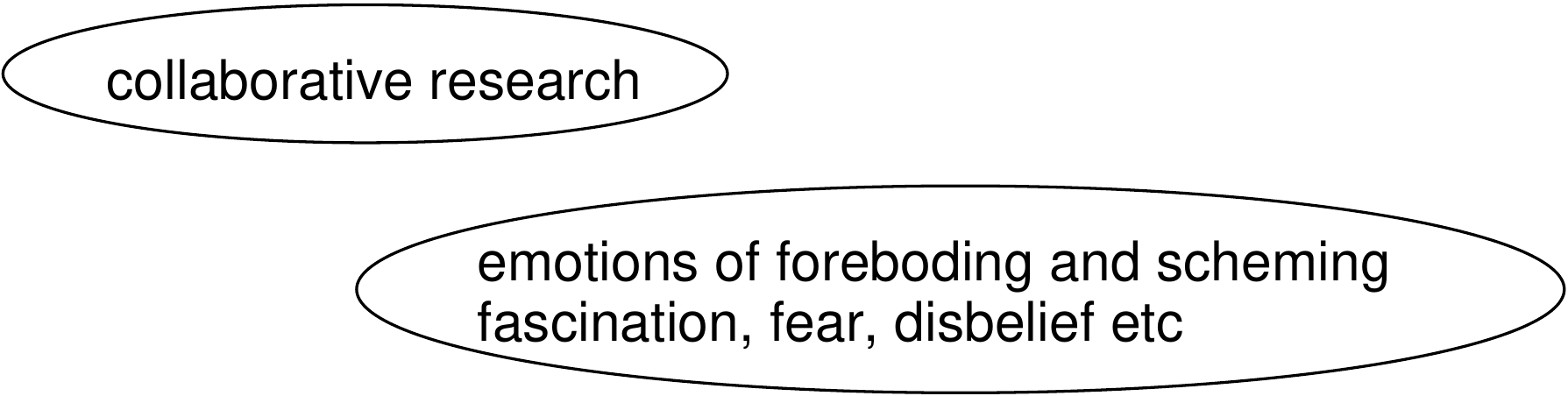}
\caption{\small The proto-emergent concepts from the novel Slogans (2 regions, from 4 links in 90 hubs).
The lack of clear subject matter is indicative of a meandering story with many themes, but
story does indeed concern research and scheming. We see a 500 page novel reduced to two short ideas,
indicating the novel is about the experience of the journey rather than the concepts induced by reading it.\label{hub_slogans}}
\end{center}
\end{figure}

With the same context ratio choices, for the novel {\em Slogans} (the
longest of the texts), remarkably only two stable regions emerged.  We
see a 500 page novel reduced to two short ideas, indicating the novel
is about the experience of the journey rather than the concepts
induced by reading it.  The themes are not persons or ideas,
but sensations: emotional characterizations of foreboding, fascination, and
anxiety, along with scheming. These are indeed central themes in the
book, and it's fascinating to see that only emotional ideas survive
the transverse contextual interferometry.  This suggests that novels
like this are more about emotional journey than a clear fact-based
reporting of subject matter (see figure \ref{hub_slogans}).

We can check the latter interpretation by comparing to {\em Moby
  Dick}---a rather better known book.  Of its six stable regions,
there is a mixture a ideas. Like another novel {\em Slogans}, these include
emotional resonances (murderous apprehension and vengeance, awe and
portending, urgency).  Some allusions to Gomorrah stand out, along
with Nantucketers (people from the region of Nantucket in the
North-Eastern seaboard of the United States). A similar emotional
emphasis is found in several other fictional sources, with the
exception of the 19th century novel {\em A Legend of Montrose} by Sir
Walter Scott, whose style is far more matter-of-fact, somewhat in the manner of Bede.

One can ask what would happen if more than one narrative were
combined.  For instance, after mixing the sea novel {\em Out of the
  Fog} with {\em Moby Dick}, the resultant map collapsed to just
two regions, one associated with emotions of bitterness and manic
thoughts, mixed in with harpoon and sailing imagery, and a second
concerning Christian experiences from environmental conditions.
The latter comes mainly from {\em Out of the Fog}, but is not alien
to {\em Moby Dick} either.

Based on these detailed analyses of the fragments, an interesting
pattern seems to emerge concerning the different kinds of narratives: a
difference between stories that are fact-based texts and stories that are
fictional tales.
\begin{itemize}
\item Non-fiction works generally led to more distinct regions, formed from
static and proper named objects, though many were fragile and were merged
spuriously as the length of the text increased.

\item In fictional tellings, the regions were fewer in number and the phrases expressed
within them concerned emotional reactions to events rather than proper names and
objects that typically characterize textbooks.
\end{itemize}
The data suggest that the purpose of fiction is not to convey facts but
rather to convey emotions: a rather different form of narrative. The
inefficiency of that approach in terms of hub separation might even
indicate why science abhors such reactions as part of its writing.
Emotional linkage muddles together factual concepts---surely a lesson
for new agencies. Although this requires a much more extensive study
to confirm or refute, it offers a tantalizing new hypothesis based on the
spacetime hypothesis.

We know that silos formed from single narratives tend to remain
isolated due to their relatively specialized use of language phrasing
(paper 1).  So one might expect fictional sources to overlap more, but
with greater entropy---creating noise rather than clarity.  Silos have
an important function in separation of concepts.  The importance of
this point should not be glossed over.  Most approaches to reasoning
are based on discrimination of increasingly specific ideas (a form of
reductionism).'  However, if concepts actually are formed from
accretion of context, this implies that there's a limit to the
robustness of logic as a reasoning system. A logic is only as good as
the concepts it starts with. If several concepts becomes merged
inseparably too early in a process of reasoning, the resultant logic
will forever be altered. Adaptive sampling, in which the effective
value of the dimensionless context ratio $\nu$ is varied, could
perhaps optimize this and make it sharper.

\subsection{Percolation catastrophe}

A looming question about the foregoing is: if memories keep getting
accumulated, wouldn't an entire network of concepts eventually
percolate, and all concepts collapse into a single pool of entropic
muddle (a version of the grey goo hypothesis)? The risk of that is low
provided links remain sparse, so as long as memory is poorly utilized
meanings should be preservable, but there is always a risk of concepts
being merged into muddle unless links and memories eventually fade to
keep a natural separation. The consequence of this is that concepts
may change over time---not only in more refined understanding, but
also the opposite. Confusion can also arise from an inability to
separate cumulative regions.

Siloed concepts have to remain independent to some degree---but how is
that line drawn? Too little support from aggregation and the world is
merely an inventory of atomic elements. If the chemistry for combining
elements is too easy or too lax, all elements would simply clump
together in a maximum entropy `grey goo'.  So far most of this study
of the Spacetime Hypothesis has been based on what is promised from
source.  Promise Theory suggests that how selections are made are at
least of equal importance\footnote{Although it took a long time to
  discover, we now know from epigenetics that realtime selection plays
  as important a role as evolutionary selection. That aspect has yet
  to be explored here in any detail\cite{epigenetics}.}.

The arbitrary scale of hub size, based on the importance horizon must
enter the discussion here. In programming an algorithm, it's all too
easy to hard code assumptions about scale that end up determining the
behaviour, as well as introducing catastrophic behavioural
transitions\footnote{This difficulty is well hidden in probabilistic
  methods, where scales are deliberately concealed by normalization in
  order to eliminate the kind of scale dependence the spoils nice
  distributions. It's threshold scales all the way down\cite{catastrophe}.}.

The choice of a fixed sampling density (of about one part in 200) is a
scale that must play a role in the outcome of these experiments. Thus
far, the effect of varying this remains to be investigated in detail.
One might expect that a higher density of sampling would yield more
accurate narratives in the input language, but not necessarily in the
effective concept language. Indeed, too high a density relative to the
proximity horizon could lead to a breakdown of separation between
concepts. In a problem with so many scales, clearly much more work is needed
to understand the various couplings.

\section{Summary and remarks}

This study extends the analysis of the so-called Spacetime Hypothesis
about cognition.  It contends that the origin and organization of what
we understand as concepts must ultimately lie in information encoded
within spacetime scales, perceived and encoded into memory by a
sensory process of an observer (called the `cognitive agent').
Space and time (as process) together imply causal narrative (time) as well as lateral
association (space).  Recall and recombination would be the mechanism
for further narrative, including what we consider to be `reasoning': a
form of combinatoric storytelling based on the spacetime constraints
implicit in memory encoding.

From the scales inherent in input patterns, boundary markers, and the
clock ticks of the sampling process, an agent can discern a
coordinatized view of its own proper timeline, exposing invariant
features that eventually acquire semantics by learning, aggregating,
and partial ordering of pattern fragments.  Locality plays a role even
in knowledge curation.  If the results of paper 1, on fractionation
and summarization, were surprisingly effective, then the results of
this sequel seem even more astonishing: given a collection of
extracted fragments, relatively simple algorithms can assemble them
into a graph, without sophisticated knowledge of language, or
descriptive logics. Only the four spacetime semantic types are needed
to distinguish time, space, descriptive scalar properties, and finally
to measure compound similarity, thus yielding a structure that
captures and summarizes narratives in the process. That structure is
basically identical to molecular chemistry, with no magical principles.

Preliminary results, on samples of data taken from books and articles,
show that the hypothesis is plausible and quite intriguing---it can't
be unequivocally confirmed, far less proven, but certainly it can't be
ruled out, but surely warrants further study on a bigger corpus. There is some
difficulty in presenting the argument using natural language as a data
source: there are pros and cons in deriving one language from another.
In some ways, the manifesto is similar to that of cognitive grammar\cite{langacker1}.

The study has been based on text analysis, for convenience, as other
significant bodies of data are hard to come by. However, the approach
is general and could be adapted to other data sources such as
quantitative time-series and process logs found in monitoring systems
of all kinds. Some changes would be needed to filter out a background
of repeating patterns, and high level of junk.  The task seems similar
to bioinformatic analysis in this respect.

Any approach to cognition is bound to have a lot of moving parts,
making presentation a challenge.  The technical approach here consists
of mining features from a serial data landscape, somewhat analogous to DNA
sequencing or crude oil refinement, in order to find the implicit
alphabet of the input language.  Samples of these alphabetic fragments
in dissolution form `context', which can then used to label events.  This method
leads to a hierarchy of implicit `containment' that mimics
generalization.  The evolution of context allows transverse overlap
between events and narratives, which defines a countable degree of
similarity by which concepts may be joined up to extend and refine
`memory regions' with sufficient stability to represent generalized
concepts over time. The linear input stream is transformed into a
multi-dimensional graph, based on the four spacetime semantic
types\cite{spacetime3,cognitive}. In future work, it would be interesting to compare this work with
other studies that seem to overlap on key principles\cite{hawkins,hawkins2,sdm1,sdm2}.
Moreover, it would be interesting to compare the geometry of the 
fragment mesh with the effective geometry of Artificial Neural Networks that
accomplish similar feats. It's not impossible that this deterministic
and causal generative process might be contribute to an explanation for the
behaviours of ANN.

The convergence of concepts into a knowledge representation is not a
smooth process---it's more like a random walk, with potentially
catastrophic changes\cite{catastrophe}. This is likely an artifact of the discreteness
of small scale `digital' separations used in natural language.  This
may be why stories about focused and curated knowledge have to evolve in
tandem with cognition: because without the structural and scale limitations of sensing,
cognition and memory encoding would cope poorly with the arbitrary
input. Unless one limits the horizon of interconnectedness, concepts
actually coalesce into a high state of entropy---and the ability to
discriminate one concept from another is lost.  Knowledge retains
coherence only within certain boundaries---understanding more about
that boundedness will be necessary in future.  Knowledge management
is, in a sense, spacetime boundary management.
By deliberately ignoring linguistics, 
data are effectively made noisier than they need to be. The more
semantics that are established, the easier it is to discriminate the
parts of the input. This we take for granted, naturally, in human
communication (at least until we try to learn a foreign tongue). This
illustrates an evolutionary pressure for a limited grammatical
recognition to emerge\footnote{In the context of Chomsky's famously
  controversial universal grammar proposal, this lends weight to the
  naturalness of the proposal, in a limited sense, but doesn't really
  confirm it or rule it out. Most likely, there are a few basic
  processing tricks that are hard coded by adaptation, which some
  authors refer to as a `propensity' for language, but far from a
  `hard coded grammar' in the explicit and literal sense that some
  attribute to it. What this work shows is that such abilities are not
  specific to language as we understand it---they would be needed for
  any sensory stream, even with specially adapted preprocessors such
  as eyes and ears. The scaling arguments are powerful and far reaching.}.
An interesting category theoretical view of concepts has been postulated
in \cite{bolt2017interacting}, but this relies on grammar for its relational
structure, but how that grammar comes about from fragments is taken for
granted, and seems to be the more interesting issue to be reckoned.
Interestingly, the authors consider sentence spaces as paths through space and time.

In summary, the structure of cognition, in this model, is rather like
the structure of form in genetics\cite{cavalli2000genes}. The
effective measure of distance between concept fragments is not an
abstract metric distance based on Euclidean embeddings in a
probability space based on a pre-identified basis\cite{feldman1}, but
rather a simple overlap of similar fragments, linked by promise (+)
and receptor (-) that fit together lock-and-key.  Distance is
analogous to the mutual information between
messages\cite{shannon1,cover1,durbin1,baldi1,nei1}. In that respect, this approach
shows that human narrative can be analyzed using the same techniques
used for DNA and other forms of chemical spectroscopy.  This is
effective because fragments are effectively invariants of a given
language.

Ultimately this work is a straightforward application of scaling
theory to discrete processes, so dimensionless scaling ratios occur as
per the Buckingham-Pi theorem.  The sensitivity of results to the
scaling ratios suggests that a well adapted organism would be able to
alter its scaling preferences in real time to maximize its clarity of
thought.  In many ways, this account of cognition is comforting.
Perhaps we don't need any mysterious new science to piece together the
underlying story of reasoning and situation awareness---only a deeper
understanding of scales.  We start with time (arrival of events), and
we invoke space by a process of discrimination. Fragments
self-organize under the constraints of external context (boundary
conditions). Scales play a central role\footnote{For all the attention
  afforded to scale-free properties in complexity studies, the scale
  free nature appear to be as much an artifact of the methodology
  rather as a property of a given process.  We seek scale invariance
  precisely because it neutralizes the semantics of scale, but scale
  plays a crucial role in spacetime measurement, i.e. in sensory
  sampling.}.  On the other hand, the study poses as many questions as
it answers: given the fragility of concepts in a sparse graph
representation, how are we humans so effective at compartmentalizing
knowledge? The work suggests that a constant rate of forgetting plays
a role, but what isn't clear yet is how so many concepts remain
sufficiently separated, without merging into a grey entropic fog. The
discreteness of the input fragments probably plays a role too.

The study only scratches the surface on what could lie beneath the
simple set of principles in the Spacetime Hypothesis.  Perhaps the
most pertinent thing we can say is: if the hypothesis does indeed play
a role in cognition, then understanding how multiple scales enable
ideas from concepts to generalizations is not beyond the reach of
human comprehension.

\bibliographystyle{unsrt}
\bibliography{spacetime}

\end{document}